\documentclass[sigconf]{acmart}

\usepackage{times}
\usepackage{latexsym}
\usepackage{booktabs}
\usepackage{graphicx}
\usepackage{subcaption}
\usepackage{xcolor}
\usepackage{multirow}
\usepackage{enumitem}
\usepackage{tikz}
\usepackage{collcell}
\usepackage{color,soul}
\usepackage{fontenc}
\usepackage{enumitem}
\usepackage{amsmath}
\usepackage{framed}
\AtBeginDocument{%
  \providecommand\BibTeX{{%
    \normalfont B\kern-0.5em{\scshape i\kern-0.25em b}\kern-0.8em\TeX}}}

\setcopyright{acmcopyright}
\copyrightyear{2018}
\acmYear{2018}
\acmDOI{XXXXXXX.XXXXXXX}

\acmConference[Conference acronym 'XX]{Make sure to enter the correct
  conference title from your rights confirmation emai}{June 03--05,
  2018}{Woodstock, NY}
\acmPrice{15.00}
\acmISBN{978-1-4503-XXXX-X/18/06}

\acmSubmissionID{901}

\begin{document}

\title{Modeling Tag Prediction based on Question Tagging Behavior Analysis of CommunityQA Platform Users}

\author{Kuntal Kumar Pal}
\authornote{Work done during the summer internship 2022 at Microsoft Research}
\email{kkpal@asu.edu}
\orcid{1234-5678-9012}
\affiliation{%
 \institution{Arizona State University}
 \city{Tempe}
 \state{Arizona}
 \country{United States}}

\author{Michael Gamon, Nirupama Chandrasekaran, Silviu Cucerzan}
\affiliation{%
 \institution{Microsoft Research}
 \city{Redmond}
 \state{Washington}
 \country{United States}}


\newcommand{\SE}{StackExchange}
\newcommand{\mask}{$\langle$mask$\rangle$}
\newcommand{\maskref}{$\langle$maskref$\rangle$}
\newcommand{\tagsep}{$\langle$tagsep$\rangle$}
\newcommand{\domain}[1]{\textsl{\textmd {#1}}}

\newcommand\blue[1]{\textcolor{blue}{#1}}
\newcommand{\silviu}[1]{\textcolor{blue}{\textbf{[#1]}}} 
\newcommand\green[1]{\textcolor{green}{#1}}
\newcommand{\mgamon}[1]{\textcolor{teal}{\textbf{[#1]}}} 
\newcommand{\kuntal}[1]{\textcolor{orange}{\textbf{[#1]}}} 
\newcommand{\niru}[1]{\textcolor{olive}{\textbf{[#1]}}} 
\newcommand\violet[1]{\textcolor{violet}{#1}}
\newcommand\cyan[1]{\textcolor{cyan}{#1}}
\newcommand{\red}[1]{\textcolor{red}{#1}}

\begin{abstract}
 In community question-answering platforms, tags play essential roles in effective information organization and retrieval, better  question routing, faster response to questions, and assessment of topic popularity. Hence, automatic assistance for predicting and suggesting tags for posts is of high utility to users of such platforms. To develop better tag prediction across diverse communities and domains, we performed a thorough analysis of users' tagging behavior in 17 \SE\space communities. We found various common inherent properties of this behavior on those diverse domains. We used the  findings to develop a flexible neural tag prediction architecture, which predicts both popular tags and more granular tags for each question. Our extensive experiments and obtained performance show the effectiveness of our model.
\end{abstract}





\maketitle

\section{Introduction}

Community Question Answering (CQA) platforms have become a very important online source of information for Web users.
On these platforms, information seeking takes the form of questions and answers in communities formed around common domains of interest.
\SE, Quora,
AnswerBag, Question2Answer, 
Reddit\footnote{stackexchange.com, quora.com, answerbag.com, question2answer.org, reddit.com} and Biostars \cite{parnell2011biostar} are some of the most
popular public CQA platforms. 
Many enterprise entities offer similar private platforms for their employees.
These communities have amassed over time large online information repositories, with high numbers of daily active users.
Thus, 
there is a 
need to organize and retrieve information efficiently,
as well as to 
facilitate question routing to interested and qualified experts in order to provide a seamless user experience and interaction.
Semantic tagging of questions plays 
an important role in this context.


Most 
CQA platforms require users to assign 
\textit{tags} to their questions. Tags are keywords representative of the topics covered by those questions.
They 
help communities to (1) \textit{categorize and organize information} (2) \textit{retrieve existing answers} for users looking for information, which in turn reduces duplicate question creation (3) \textit{route questions to topic experts} which improves 
query response time 
and answer quality
(4) \textit{provide tag-based notifications}, which allow knowledgeable community members to answer questions in their areas of expertise and gain reputation
(5) \textit{assess the popularity} of various areas and topics in the targeted domain.



Asking users to annotate their questions with
tags without providing adequate support 
poses several 
challenges, in particular with respect to novice users and to the 
lack of knowledge 
about tag usage in a community, which may lead to the creation of various tags with the same meaning, as well as different orthographic forms of those tags. 
This makes question routing difficult (for tag-based subscription platforms), delays response time, and leads to poor 
information organization. In turn, addressing these issues would require community administrators to constantly work on identifying and merging near-duplicate tags. 
Additionally, lack of support in suggesting adequate tags may inhibit novice users from asking questions and/or lead to questions being mistagged and not answered. 
These challenges may become more severe in enterprise CQA platforms due to community size and topic sparsity.
Against this background,
 tag-prediction becomes an extremely important while
 challenging task for both public and private CQA platforms. In this investigation,
 our first goal was to 
 \textit{understand the commonalities
 of the tagging behaviors} of 
 users through a large scale analysis of 17 diverse domains in 
 \SE\space (Section \ref{sec:tag_behavior_analysis}). 
 Our 
 analysis revealed that while these 
 domains are quite diverse in terms of volume of questions, users and tags, they share 
 common 
 distributional properties for tag and tag pair usage.
 Also, there is a large lexical
 overlap between the tags and user texts in every domain. Post coverage of tags is high in all domains. Tags also show positional stability and tag pairs show particular ordering preferences forming a soft hierarchy among 
 tags. 
 

\begin{figure*}[!htb]
    \centering
    \includegraphics[scale=0.065]{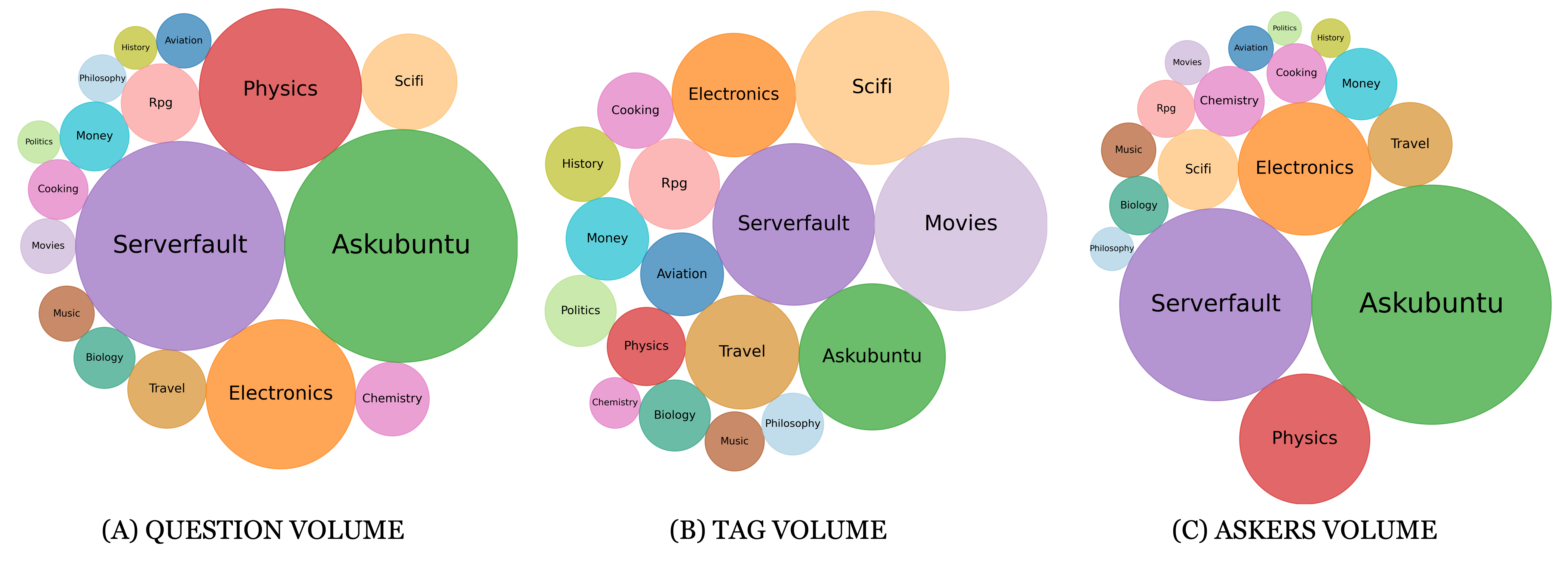}
    \caption{Community Diversity in terms of Volume}
    \label{fig:diversity}
\end{figure*}

 We incorporate the findings to develop a neural model with two tag-prediction heads - one 
 trained to predict existing
 popular tags such as the name of important topics in a domain (e.g. "harry-potter", doctor-who", and "star-wars" in the \domain{scifi} domain) and frequently-used \textit{meta-tags} (e.g. "video-games", "books", and "short-stories" in \domain{scifi})
 and another one
 generate 
 \textit{finer-grained tags}, which may have been used rarely on previous questions or are new. 
 Typically, the former category of tags
 represent the main topic area
 of a question while the latter 
 help in further scoping down and
 clarifying it.
 Both types of tags are equally important in identifying the question and hence it is necessary for the tag prediction systems to not only predict the main generic tags but also the refined ones.
 
 Our experiments show that the proposed approach
significantly outperforms baseline methods in prediction of both generic tags and finer-grained tags. We also investigate and 
show the effect of reducing the pre-defined vocabulary size,
as well as the contributions of each prediction head.
Our main contributions in this work are:

\begin{itemize}[leftmargin=*]
    \item 
    We present an in-depth analysis of the tagging behaviors of the users of a CQA  platform (\SE) on 17 diverse domains. We present our findings of question tag analysis  across four dimensions: tag space, tag co-occurrence, tag pair ordering, and tag positional stability.
    \item We propose a tag prediction architecture for both predicting popular tags from a pre-defined vocabulary and generating refined tags not present in the vocabulary. 
    \item We perform comprehensive experiments on the 17 domains and show effects of each model component under various experimental settings.
\end{itemize}

\section{Dataset Preparation}

We collected data from 17 
communities of \SE\space that correspond to a diverse set of domains. 
We use the \SE\space data dumps\footnote{https://archive.org/details/stackexchange\_20210301}  (2021-03-01) for  our analysis and model. 
We find that the Post.xml file is sufficient for our tag analysis and predictions.
We only consider the posts from the dataset which are either questions or answers (\textit{PostTypeId}) for our analysis. We  reject posts with no owners (\textit{OwnerUserId}, \textit{OwnerDisplayName}).
As imposed by \SE,
the minimum and maximum number of tags assigned to 
each posts are   one and five respectively 
and all the posts in this data set are dated prior to 
March, 2021. We chose several 
domains from each of the 
following 
\SE\space categories\footnote{https://stackexchange.com/sites\#}: Technology, Culture \& recreation, Life \& arts, Science and Professional. Each selected
domain has at least a decade of posts.
We do not include the \domain{stackoverflow} domain because of its enormous volume and also a random sample set might not be representative of the full data of this domain. Hence we consider \domain{askUbuntu} which is also a representative community of the \textit{Technology} domain.

\section{Tagging Behavior Analysis}
\label{sec:tag_behavior_analysis}
To understand the user behavior of question tagging and to identify the inherent commonalities, we analyze ten years of data
from these 17 
domains.

\noindent
\textbf{Mathematical Notation:} Without loss of generality, let $D$ denote one of the domains (out of 17) being investigated, $P$ the set of posts in the data for this domain, and $T=\{t_1, t_2,\dots,t_{|T|}\}$ the set of all tags used in domain $D$.
Each post $p_j\in P$ has associated a sequence of tags $S(p_j)=\big(t_{(1)},t_{(2)},\dots,t_{(l)}\big)$, $1\leq l\leq 5$, where $t_{(i)}$ denotes the tag at position $i$ in that sequence. We employ parentheses to distinguish between the positional information of a tag in a sequence and the indexes that identify elements $t_i$ of the tag set $T$ observed for domain $D$.


\begin{figure*}
    \centering
    \includegraphics[width=\textwidth,height=4.5cm]{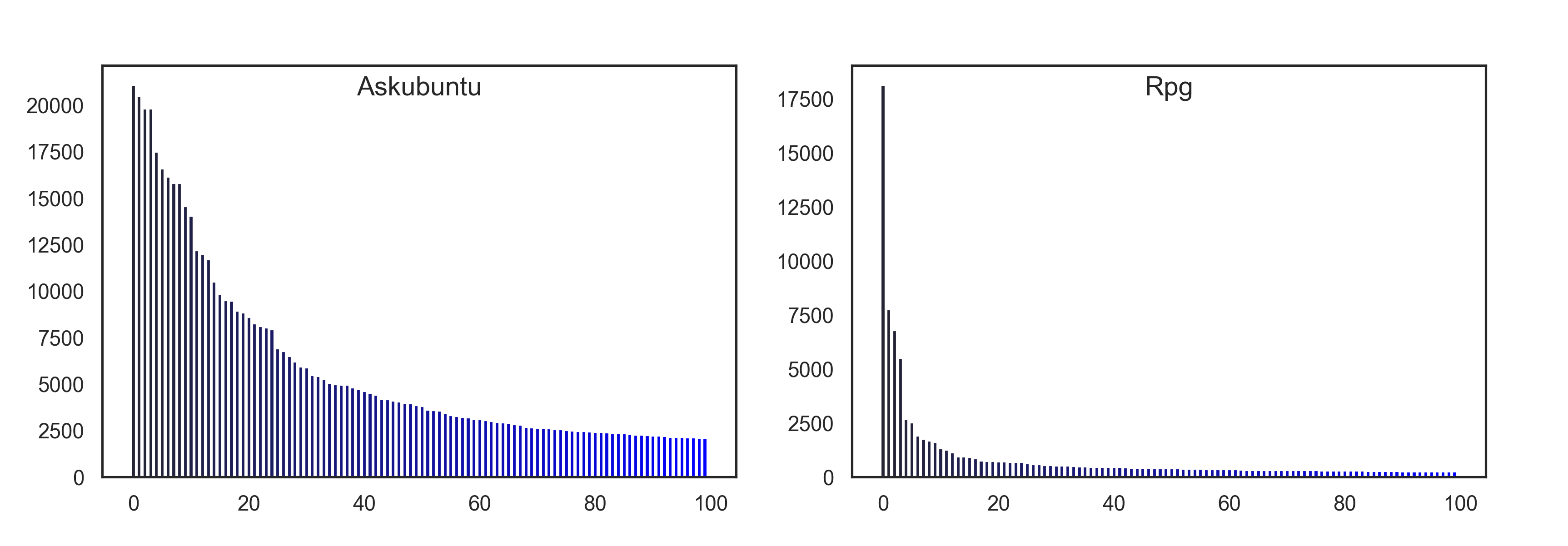}
    \caption{Tag Distribution Patterns,Y-Axis: \# posts the tag appear in, X-Axis: Top-100 Tags}
    \label{fig:tag_distrib_2}
\end{figure*}

\subsection{Community Diversity}
\label{sec:comm_div}
We observed a high degree of variability for the selected domains 
in terms of \textit{Question Volume}, \textit{Tag Space} and \textit{Asker Volume}. Figure \ref{fig:diversity} shows a 
visual comparison of this variability, while 
Table \ref{tab:data_stats} shows general statistics for each domain.
In terms of the amount of information created over a decade, \textit{only four} domains have over 100K posted questions while the domains \domain{politics} and \domain{history} 
have merely 12K. If we consider the number of unique tags (\#T) created, 
the domain \domain{movies} ranks
highest, as new movie titles are added to the tag set on weekly basis. 
To quantify tag 
re-use in each domain, we define \textit{post-per-Tag (PPT)} as the number of posts available for one tag. We also observe that \domain{physics}, \domain{askubuntu}, and \domain{chemistry} are domains with the most tag-reuse (PPT $>$ 100) while  \domain{movies} domain (PPT $<$ 5) shows frequent new tags.
The number of posts having views over 100 (V$>$100) can be used to infer the popularity of posts in each domain. From the average number of tags (AvgT) per post, we can infer the need for detailed tagging in each domain.
In \domain{travel}, \domain{physics}, and \domain{money}, AvgT $>$ 3 indicates users feel the need to assign more than 3 tags to clarify their questions. Also, the movie domain has the least AvgT (2.09), showing that only two tags on average are sufficient. 
Some domains like \domain{aviation}, \domain{philosophy}, \domain{history}, \domain{movies}, \domain{politics} are not popular (\#A $<$ 10K in a decade).
More statistics are in Appendix Table \ref{app:tab_more_stats}.

\begin{table}[t]
\caption{Community Diversity. V:Views, PPT:Posts/Tag, QPA:\#Q/\#A,  AvgT: Average \#T per Q, \#T: Unique Tags}
\label{tab:data_stats}
\resizebox{\linewidth}{!}{%
\begin{tabular}{@{}l|rrrrrrr@{}}
\toprule
\multicolumn{1}{c}{Domain} &
  \multicolumn{1}{c}{\#Q} &
  \multicolumn{1}{c}{\#T} &
  \multicolumn{1}{c}{PPT} &
  \multicolumn{1}{c}{AvgT} &
  \multicolumn{1}{c}{V$>$100} &
  \multicolumn{1}{c}{\#A} &
  \multicolumn{1}{c}{QPA}\\ \midrule
 askubuntu & 371800 & 3121 & 119.13 & 2.78 & 1093 & 201912 & 1.84 \\
aviation & 20345 & 1002 & 20.30 & 2.56 & 12 & 7066 & 2.88 \\
biology & 25671 & 739 & 34.74 & 2.58 & 11 & 12089 & 2.12 \\
chemistry & 37476 & 375 & 99.94 & 2.37 & 7 & 17202 & 2.18 \\
cooking & 24513 & 833 & 29.43 & 2.30 & 13 & 12413 & 1.97 \\
electronics & 152980 & 2226 & 68.72 & 2.77 & 36 & 61869 & 2.47 \\
history & 12562 & 813 & 15.45 & 2.84 & 19 & 5296 & 2.37 \\
money & 32648 & 995 & 32.81 & 3.11 & 37 & 18010 & 1.81 \\
movies & 20749 & 4348 & 4.77 & 2.09 & 30 & 6931 & 2.99 \\
music & 20925 & 512 & 40.87 & 2.52 & 5 & 10447 & 2.00 \\
philosophy & 15624 & 559 & 27.95 & 2.40 & 6 & 6640 & 2.35 \\
physics & 180166 & 893 & 201.75 & 3.17 & 131 & 59774 & 3.01 \\
politics & 12416 & 739 & 16.80 & 2.90 & 27 & 3970 & 3.13 \\
rpg & 42693 & 1195 & 35.73 & 2.91 & 56 & 11541 & 3.70 \\
scifi & 62987 & 3433 & 18.35 & 2.25 & 153 & 22717 & 2.77 \\
serverfault & 299895 & 3814 & 78.63 & 2.90 & 327 & 130214 & 2.30 \\
travel & 42201 & 1891 & 22.32 & 3.28 & 42 & 24895 & 1.70 \\
\bottomrule
\end{tabular}%
}
\end{table}

\subsection{Tag-Space Analysis}
We analyzed each domain's tag spaces into (1) General Tag Statistics (2) Tag Distributions (3) Tag-Post Coverage (4) Tag-Post Overlap.

\noindent
\textbf{General Tag Statistics:}
The shortest tag in every domain is merely 1-3 characters long (\textit{c}, \textit{air}, \textit{3g}) while the longest tag is 22-35 characters long (\textit{valerian-city-of-a-thousand-planets}, \textit{neurodegenerative-disorders}). \domain{askubuntu} has the lowest average tag length (8.17) while \domain{movies} has the highest (13.66). We believe that the tags in \domain{askubuntu} are short technical terms of a subtopic but movie names tend to be quite long in comparison and are often used as a part of a tag in the movie domain. 
Table \ref{tab:tag_word_length} shows the distribution based on the number of words of the tags. With the exception of \domain{movies}, \domain{rpg}, and \domain{scifi} the majority of tags in all the domains consist of three or fewer words.
The shortest and longest tags for each domain are presented in Appendix Table \ref{tab:app_tag_stats}.

\begin{table}
\caption{Tag \% based on the Number of Words in the Tag}
\label{tab:tag_word_length}
\tiny
\centering
\resizebox{0.95\columnwidth}{!}{%
\begin{tabular}{@{}l|rrrrrr@{}}
\toprule
Domain & 1 & 2 & 3 & 4 & 5 & \textgreater{}5 \\ \midrule
askubuntu & 80.83 & 18.73 & 0.37 & 0.07 & 0 & 0  \\
aviation & 49.74 & 43.86 & 6.34 & 0.05 & 0 & 0  \\
biology & 69.30 & 29.95 & 0.75 & 0 & 0 & 0  \\
chemistry & 47.17 & 50.36 & 2.31 & 0.16 & 0 & 0  \\
cooking & 78.53 & 21.11 & 0.36 & 0.01 & 0 & 0  \\
electronics & 74.23 & 23.9 & 1.33 & 0.54 & 0 & 0  \\
history & 56.86 & 36.1 & 7.01 & 0.03 & 0 & 0  \\
money & 50.00 & 45.51 & 4.05 & 0.45 & 0 & 0  \\
movies & 32.81 & 41.58 & 16.32 & 5.61 & 2.57 & 1.1  \\
music & 77.74 & 21.07 & 1.17 & 0.02 & 0 & 0  \\
philosophy & 69.42 & 14.02 & 16.29 & 0.27 & 0.01 & 0  \\
physics & 41.37 & 49.31 & 9.02 & 0.3 & 0 & 0  \\
politics & 51.26 & 45.05 & 3.59 & 0.08 & 0.02 & 0  \\
rpg & 42.43 & 51.39 & 4.82 & 1.11 & 0.16 & 0.09  \\
scifi & 31.04 & 49.23 & 13.19 & 3.23 & 2.1 & 1.2  \\
serverfault & 67.91 & 23.09 & 7.62 & 1.32 & 0.06 & 0  \\
travel & 65.78 & 26.5 & 6.87 & 0.85 & 0 & 0  \\
\bottomrule
\end{tabular}%
}
\end{table}

\noindent
\textbf{Tag Distributions:} 
\textit{There is a  long tail in the distribution of tags in every domain} (Figure \ref{fig:tag_distrib_2}). 
We observe that (1) most larger domains where the tag re-use is high, have smoother tag distributions like \domain{askubuntu},  \domain{electronics}, \domain{biology} and (2) for some smaller domains like \domain{scifi}, \domain{movies}, \domain{rpg}, the most frequent tag dominates the distribution.
The rest of the distributions are shown in the Appendix Figure \ref{fig:app_tag_distrib}. Also, Table \ref{tab:tag_post_coverage} shows that the 100 most frequent tags (100Tag\%) constitute a very small portion of the tag space for large domains.

\begin{table}[]
\caption{Top-n Tag's Post Coverage. \#T:\#distinct tags, 100Tag\%:Frequent 100 tag \% among whole tag-space.}
\label{tab:tag_post_coverage}
\tiny
\centering
\resizebox{0.9\columnwidth}{!}{%
\begin{tabular}{@{}l|rr|rrrr@{}}
\toprule
\multicolumn{1}{l}{Domain} &
  \multicolumn{1}{l}{\#T} &
  \multicolumn{1}{l}{100Tag\%} &
  \multicolumn{1}{l}{Top1} &
  \multicolumn{1}{l}{Top10} &
  \multicolumn{1}{l}{Top100} \\ \midrule
 askubuntu & 3121 & 3.20 & 5.67  & 40.21 & 82.68  \\
aviation & 1002 & 9.98 & 11.05 &  45.93  & 89.43  \\
biology & 739 & 13.53 & 9.22 & 55.05 & 91.76 \\
chemistry & 375 & 26.67 & 23.05 & 61.38 & 95.35 \\
cooking & 833 & 12.00 & 9.55  & 38.99  & 85.19  \\
electronics  & 2226 & 4.49 & 4.94 & 32.81 & 81.98  \\
history & 813 & 12.30 & 10.86  & 45.91  & 89.95 \\
money & 995 & 10.05 & 37.04  & 68.52  & 94.18 \\
movies & 4348 & 2.30 & 36.93 & 66.84 & 85.88 \\
music & 512 & 19.53 & 14.93  & 58.04 & 94.54  \\
philosophy & 559& 17.89 & 19.39  & 63.30  & 93.77  \\
physics & 893 & 11.20 & 12.70  & 55.10 & 91.68  \\
politics & 739 & 13.53& 46.00  & 66.41  & 94.95  \\
rpg & 1195 & 8.37 & 42.50 & 79.75  & 92.66  \\
scifi & 3433 & 2.91 & 27.86 & 70.67 & 85.04 \\
serverfault & 3814 & 2.62& 11.92 & 42.76 & 82.86  \\
travel & 1891 & 5.29 & 22.20  & 58.34  & 92.36  \\
\bottomrule
\end{tabular}%
}
\end{table}

\begin{figure*}
    \centering
    \includegraphics[width=0.8\textwidth]{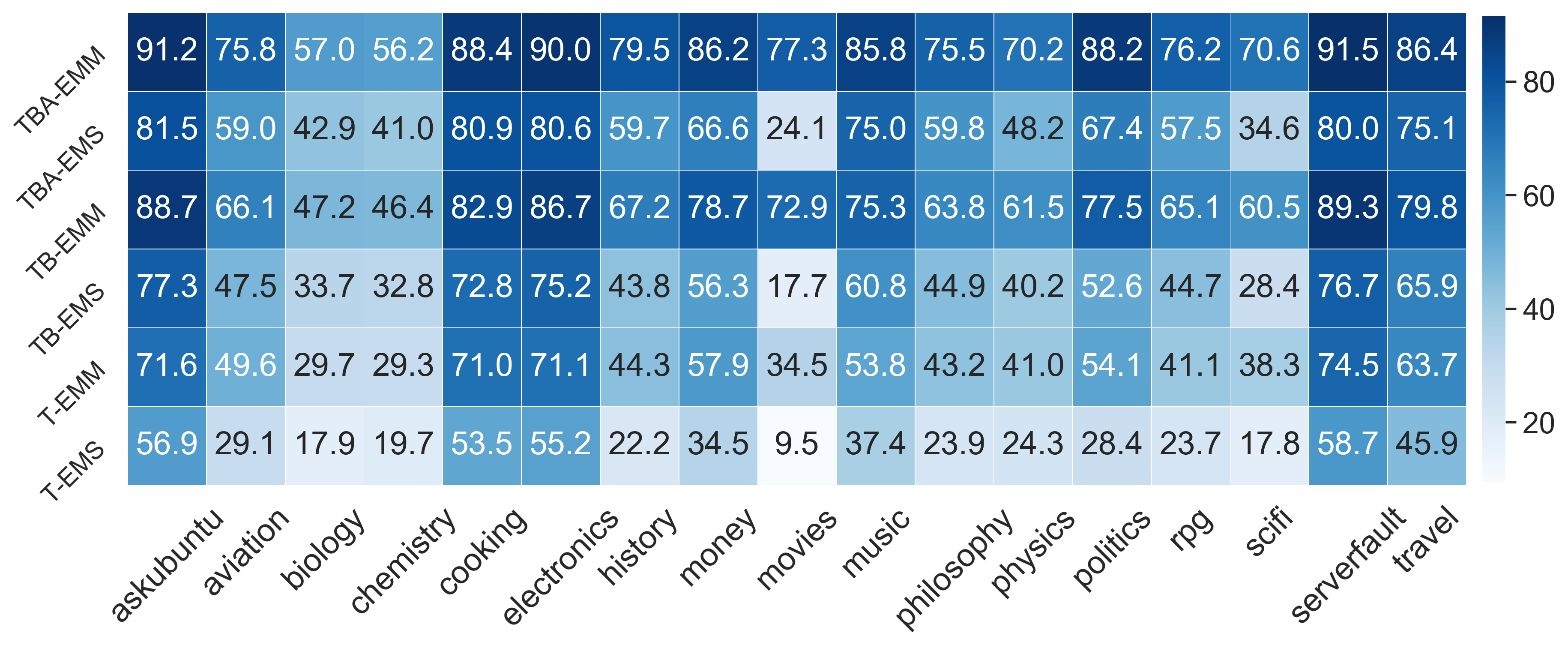}
    \caption{Tag-Post Overlap: \% posts where at least one tag appears in user texts: Title(T), Body(B), Answers(A). EMS\%: single word tag exact match, EMM\%: single \& multi-word tag exact match. }
    \label{fig:tag_post_overlap}
\end{figure*}

\noindent
\textbf{Post Coverage by Tags:} We consider a tag to cover a post if it is present in the tag sequence of the post. Table \ref{tab:tag_post_coverage} shows the percentage of total posts that can be covered by the top $n$ most frequent tags in each domain. 
We observe that the most frequent tag covers (Top1) at most 10\% of posts in \domain{electronics}, \domain{askubuntu}, \domain{cooking}, and \domain{biology} domains but more than 40\% in \domain{politics} and \domain{rpg} domains.
More than 81\% of all posts in each domain are covered by the 100 most frequent tags.

\noindent
\textbf{Tag-Post Overlap:}
\label{sec:tag_post_ovrlap}
Figure \ref{fig:tag_post_overlap} shows whether the tags appear in user contents (question-title / question-body / answers) using two metrics: (1) \textit{single worded tag exact-match (EMS)} and \textit{both single and multiple worded tag exact-match (EMM)}. 
We observe that in 8/17 domains, tags appear in more than 50\% of post titles. The movie domain has more multi-worded tags than single worded tags (9.49\% compared to 34.51\%). Two science domains - \domain{biology} and \domain{chemistry} - have the lowest  tag overlap ($<$30\%) with the question title (T-EMS). When we include the question body, we observe, in 9/17 domains, question tags appear in more than 70\% of posts. Finally, if we include every answer for each question, all the domains (except \domain{chemistry} and \domain{biology})  have their tags appear in more than 70\% of the posts. The three larger domains (\domain{askubuntu}, \domain{serverfault}, and \domain{electronics}) have more than 90\% overlap. The overlap is lowest (56\%) for the \domain{chemistry} and \domain{biology} domains.





\begin{table}
\caption{Tag Pairs Post Coverage : \% posts covered by top-k tag pairs. Single: \% of posts with single tag}
\label{tab:tagpair_post_coverage}
\resizebox{0.95\columnwidth}{!}{%
\begin{tabular}{l|rrrrrr|r}
\toprule
    Domain & Top-1 & Top-3 & Top-5 & Top-10 & Top-50 & Top-100 & Single \\
    \midrule
    askubuntu & 1.57 & 2.89 & 5.33 & 9.43 & 17.97 & 23.45 & 17.70 \\
    aviation & 2.05 & 3.49 & 4.78 & 6.99 & 17.00 & 23.81 & 19.27 \\
    biology & 2.85 & 4.90 & 7.41 & 11.39 & 25.85 & 33.34 & 20.67 \\
    chemistry & 4.33 & 7.62 & 9.99 & 14.56 & 29.82 & 36.95 & 23.89 \\
    cooking & 1.60 & 3.45 & 4.34 & 5.89 & 13.51 & 18.54 & 25.81 \\
    electronics & 0.76 & 2.16 & 3.20 & 5.08 & 13.03 & 18.62 & 18.31 \\
    history & 2.37 & 4.86 & 6.09 & 9.93 & 20.97 & 27.58 & 15.34 \\
    money & \blue{10.39} & \blue{17.13} & \blue{18.52} & \blue{24.16} & \blue{39.92} & \blue{46.49} & 10.51 \\
    movies & 2.50 & 6.28 & 7.81 & 10.93 & 20.29 & 25.30 & 21.98 \\
    music & 2.48 & 5.32 & 7.56 & 13.52 & \blue{31.20} & 38.17 & 20.49 \\
    philosophy & 1.74 & 4.97 & 7.32 & 11.08 & 26.54 & 33.79 & 27.85 \\
    physics & 2.32 & 5.29 & 7.10 & 11.07 & 28.54 & 37.46 & 11.39 \\
    politics & 4.59 & \blue{11.98} & \blue{17.60} & \blue{27.24} & \blue{43.27} & \blue{49.24} & 10.40 \\
    rpg & \blue{12.48} & \blue{17.23} & \blue{22.27} & \blue{28.13} & \blue{43.54} & \blue{52.73} & 9.96 \\
    scifi & 5.58 & \blue{12.09} & \blue{17.92} & \blue{26.12} & \blue{43.57} & \blue{49.29} & 25.86 \\
    serverfault & 1.09 & 2.84 & 4.16 & 6.23 & 16.07 & 22.29 & 13.03 \\
    travel & 5.17 & \blue{12.01} & \blue{14.64} & 18.04 & \blue{31.01} & 38.43 & 6.45 \\
 \bottomrule
\end{tabular}
}
\end{table}

\subsection{Tag Co-Occurrence Analysis}
\label{sec:hierarchy}
For a post $p_k$, we define tag co-occurrence $C_{ij} = \{\{t_i,t_j\}:t_i,t_j\in S(p_k), t_i\neq t_j\}$ as a pair of tags $\{t_{i}, t_{j}\}$ appearing in a post together irrespective of their positions.

\noindent
\textbf{Soft Tag Hierarchy:}
From the tag co-occurrence analysis in the 17 domains, we find that \textit{there exists a soft hierarchy among the tag pairs}.
One of the tags 
indicates the main topic or area of the question and the other tag is often  fine-grained  which 
makes the question more specific. For these
examples, the second tag is a sub-category of the first: 
\textit{(baking, bread)} in \domain{cooking}, \textit{(dnd-5e, spells)} in \domain{rpg} and \textit{(aircraft-design, wing)} in \domain{aviation}. 
In the science domain, similar examples of topic-subtopic relationships  are
 \textit{(organic-chemistry, carbonyl-compounds)} in \domain{chemistry} and \textit{(hilbert-space, quantum-mechanics)} in \domain{physics}. 
 The most frequently occuring tag-pair for each domain is shown in Table \ref{tab:most_freq_cooccur}, in  Appendix Table \ref{tab:app_tag_cooccur_top5} a more comprehensive set of the top-5 most frequent pairs per domain are shown. 



\begin{figure*}
    \centering
    \includegraphics[scale=0.052]{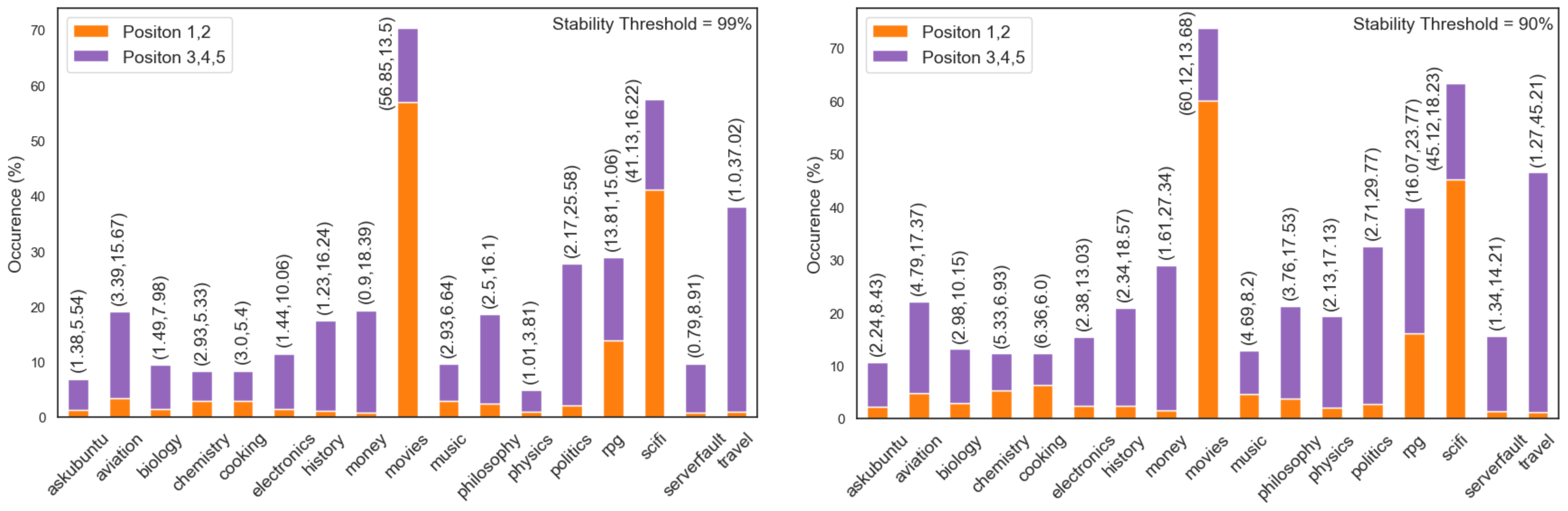}
    \caption{Tag Position Stability : $\delta=$ 99\% (Left) and $\delta=$ 90\% (Right)}
    \label{fig:tag_stability}
\end{figure*}

\begin{table}[]
\caption{Most Frequently Co-Occurring Tag-Pairs}
\label{tab:most_freq_cooccur}
\resizebox{\columnwidth}{!}{%
\begin{tabular}{@{}l|lr@{}}
\toprule
Domain & Top Pair & Post-Count \\ \midrule
       askubuntu  &  ('boot', 'grub2')  & 5845\\
aviation  &  ('aerodynamics', 'aircraft-design')  & 417\\
biology  &  ('entomology', 'species-identification')  & 731\\
chemistry  &  ('organic-chemistry', 'reaction-mechanism')  & 1621\\
cooking  &  ('baking', 'bread')  & 393\\
electronics  &  ('current', 'voltage')  & 1161\\
history  &  ('nazi-germany', 'world-war-two')  & 298\\
money  &  ('taxes', 'united-states')  & 3393\\
movies  &  ('character', 'plot-explanation')  & 518\\
music  &  ('chords', 'theory')  & 519\\
philosophy  &  ('logic', 'philosophy-of-mathematics')  & 272\\
physics  &  ('homework-and-exercises', 'newtonian-mechanics')  & 4182\\
politics  &  ('donald-trump', 'united-states')  & 570\\
rpg  &  ('dnd-5e', 'spells')  & 5330\\
scifi  &  ('short-stories', 'story-identification')  & 3514\\
serverfault  &  ('linux', 'ubuntu')  & 3261\\
travel  &  ('uk', 'visas')  & 2181\\
       \bottomrule
\end{tabular}%
}
\end{table}

\noindent
\textbf{Tag Pair Post Coverage:}
We consider a tag-pair (\{$t_i$,$t_j$\}) to cover a post if the tag-pair occurs in the sequence of tags for that post in any position.
Table \ref{tab:tagpair_post_coverage} shows the tag pair post coverage across the domains. We see around 10-20\% of posts have only a single tag. Considering the most frequent 100 pairs we can cover 18-53\% posts. Also, the most frequent tag pair can cover more than 10\% of posts in \domain{money} and \domain{rpg} domains which shows that this tag-pair is extremely essential for these two domains.

\noindent
\textbf{Tag Pair Distribution:}
On analyzing the distribution of top-50 frequently occurring tag pairs in each domain, we observe three patterns: (1) \textit{Smooth Distribution} (2) \textit{Spike in Top-1} and (3) \textit{Spikes in top few pairs}.
Larger domains (\domain{askubuntu}, \domain{serverfault}, \domain{electronics}) have smooth distributions. 
In smaller domains (\domain{movies}, \domain{scifi}, \domain{travel}) few tag pairs dominate the distributions, indicating their popularity. More Details are available in Appendix Section \ref{sec:tag_cooccur} and Figures \ref{fig:app_tag_cooccur} and  \ref{fig:tag_cooccur_3}.

\subsection{Tag Pair Ordering}
\label{sec:tag_pair_ordering}
We analyze the top-10 most frequent tag pairs in each domain to identify users' ordering preferences for tags.
For a post $p_k$, $O_{ij}=(t_{(m)},t_{(n)})$ (and $O_{ji}$) are the tag ordering for the tag pairs $t_i$ and $t_j$, where $m$ and $n$ are the positions of $t_i$ and $t_j$ respectively in the tag sequence $S(p_k)$.
We find that community users have a tendency to assign the more generic tags prior to the specific ones, for each domain by analyzing the occurrence of $O_{ij}$ and $O_{ji}$. 
For example, \textit{aircraft-design} always appears before \textit{wings} out of 221 times they appear together in \domain{aviation}, \textit{united-states} appears before \textit{income-tax}, 99.95\% of times out of 3393 times they appear in the \domain{money} domain and \textit{dnd-5e} always appears before \textit{magic-items} out of 1367 times in \domain{rpg}.
More examples are in the Appendix \ref{sec:app_tag_ordering}. 

\subsection{Tag Position Stability}
\label{sec:tag_stability}


\begin{table*}[]
\caption{Sets of five randomly picked stable tags for positions 1,2 and five for positions 3, 4, 5, respectively, across 17 domains.}
\label{tab:app_stable_eg}
\resizebox{0.95\textwidth}{!}{%
\begin{tabular}{@{}l|ll@{}}
\toprule
Domain & Position 1, 2 & Position 3, 4, 5 \\ 
\midrule
askubuntu & ['software-installation','server','community','locoteams','10.04'] & ['multiple-workstations','equalizer','speakers','workflow','flicker'] \\ 
aviation & ['air-traffic-control','radio-communications','airspace','flight-planning','faa-regulations'] & ['rotary-wing','rvsm','sfo','dash-8','special-vfr'] \\ 
biology & ['biochemistry','immunology','cell-biology','dna','molecular-biology'] & ['ribosome','binding-sites','exons','dendritic-spines','rna-interference'] \\ 
chemistry & ['crystal-structure','equilibrium','organic-chemistry','thermodynamics','inorganic-chemistry'] & ['nitro-compounds','bent-bond','phenols','organosulfur-compounds','reaction-coordinate'] \\ 
cooking & ['baking','oven','eggs','substitutions','sauce'] & ['oregano','condensed-milk','chopping','blind-baking','scottish-cuisine'] \\ 
electronics & ['arduino','motor','soldering','ethernet','avr'] & ['basic-stamp','debugwire','sinking','nxp','fuse-bits'] \\ 
history & ['20th-century','world-war-one','language','china','political-history'] & ['proof','dday','crusaders','templars','republic-of-ireland'] \\ 
money & ['investing','united-states','canada','taxes','credit-card'] & ['pension-plan','contractor','contribution','limits','debt-reduction'] \\ 
movies & ['wedding-crashers','analysis','star-wars','comedy','the-pink-panther'] & ['manichitrathazhu','chandramukhi','bhool-bhulaiyaa','clint-eastwood','for-a-few-dollars-more'] \\ 
music & ['learning','voice','theory','tuning','scales'] & ['stick-control','archeterie','instrumentation','rsi','rock-n-roll'] \\ 
philosophy & ['epistemology','philosophy-of-mathematics','ethics','existentialism','logic'] & ['dreams','plantinga','rationalism','rule-ethics','arithmetic'] \\ 
physics & ['quantum-mechanics','particle-physics','string-theory','acoustics','experimental-physics'] & ['action','faq','stability','wavefunction-collapse','coriolis-effect'] \\ 
politics & ['election','political-theory','democracy','united-kingdom','israel'] & ['first-past-the-post','checks-and-balances','redistricting','faithless-elector','puerto-rico'] \\ 
rpg & ['pathfinder-1e','dnd-3.5e','game-recommendation','dungeons-and-dragons','dogs-in-the-vineyard'] & ['feywild','group-scaling','round-robin-gming','romance','charmed'] \\ 
scifi & ['novel','vorkosigan-saga','total-recall-2070','star-trek','the-road'] & ['star-trek-data','3001-the-final-odyssey','rama-revealed','star-trek-emh','skylark-series'] \\ 
serverfault & ['sql-server','backup','sql-server-2008','raid','windows'] & ['tempdb','fakeraid','tuning','su','debian-etch'] \\ 
travel & ['loyalty-programs','transportation','public-transport','sightseeing','safety'] & ['amazon-river','amazon-jungle','singapore-airlines','sin','trans-siberian'] \\ 
\bottomrule
\end{tabular}%
}
\end{table*}

\begin{figure}
   \centering
   \includegraphics[width=\columnwidth]{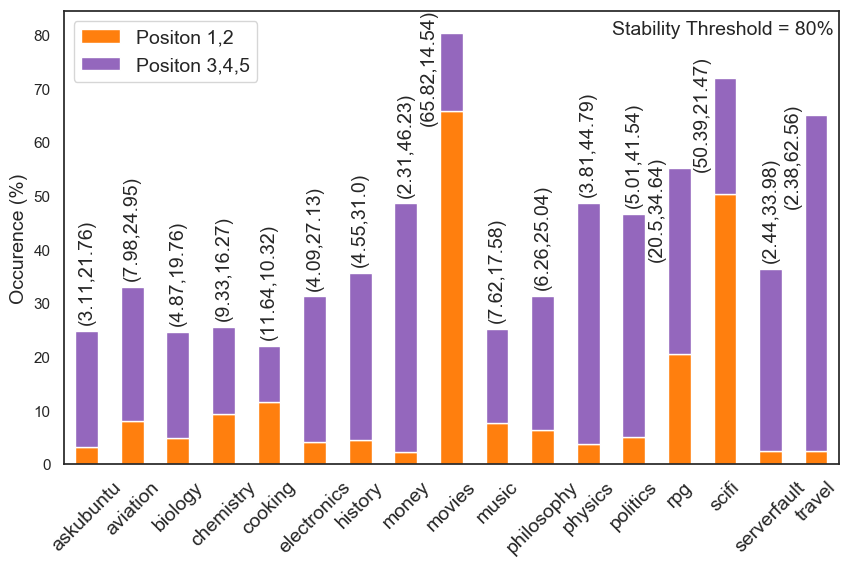}
   \caption{Tag Stability (80\%)}
   \label{fig:app_tag_stability_80}
\end{figure}

We study the positional stability of tags i.e., whether some tags frequently appear in any particular position among the five allowed by \SE.
We consider $\phi_x (t)$ as the percentage of occurrence of a tag ($t$) in any position $x$, given by,
\begin{equation}
    \phi_x(t) = \frac{c(t_{(x)})}{\sum_{k=1}^{5}{c(t_{(k)})}} \%
\end{equation}
\noindent
where $c(t_{(x)})$ denotes the count of tag $t$ in position $x$. We consider three stability thresholds ($\delta$) - 80\%, 90\%, 99\% (Figure \ref{fig:tag_stability} and \ref{fig:app_tag_stability_80}). For a tag $t$ and position $x$, $\phi_x(t) > \delta$ indicates that the tag is stable at that position. 

{
\begin{gather}
    Q_{X} = \{ t \in T: \sum_{x\in X}\phi_x(t)\geq \delta \} \\
    ST_{X} = \frac{|Q_{X}|}{|T|} \%
\end{gather}
}

\noindent
where $Q_{X}$ is the set of tags that occurs more than $\delta$ in sets of positions defined by $X$ and $ST_X$ is the percentage of tags in a domain that are stable at positions $X$.
In Figure \ref{fig:tag_stability}, (\domain{rpg} domain) for $\delta = 99\%$, we find  $ST_{1,2}=13.81$ i.e. 13.81\% of all tags in \domain{rpg} are stable in positions 1 and 2 combined, and $ST_{3,4,5}=15.06$ are stable in positions 3, 4 and 5 combined. The rest of the tags are unstable. Also, the stable tags ($Q_{3,4,5}$) appearing in positions 3, 4, and 5 are finer-grained (or refined) tags that support the stable tags present in positions 1 and 2 ($Q_{1,2}$). 

The \domain{travel} domain, has the highest number of stable tags appearing in positions 3,4, and 5 ($Q_{3,4,5}$) with $\delta=90\%, 99\%, 80\%$ threshold showing that to make a question specific more than one refined tags is needed in this domain. We neither find any conclusive evidence of this stability within positions 1 and 2 (i.e. $Q_1$ and $Q_2$), nor within positions 3, 4 and 5 (i.e. $Q_3$, $Q_4$ and $Q_5$) individually.

Table \ref{tab:app_stable_eg} shows five randomly selected examples of position-stable tags in 17 domains. These positions account for more than 99\% of the occurrences of these tags in their respective domains.

\section{Modeling Tag Prediction}
Based on the observations from our tagging behavior analysis (Section \ref{sec:tag_behavior_analysis}), we develop an automated generic tag prediction approach for CQA platforms that predicts both generic and refined tags.
The inherent commonalities in community diversity influence our decision to develop a common tag generation framework. The long tail in tag-space analysis guided us to develop a predictive-generative hybrid model. Tag co-occurrence analysis, tag-pair ordering, and tag-positional analysis on these domains led us to generate $n$ tags from a common vocabulary of popular tags at certain positions and $m$ related granular tags at the remaining positions.

\subsection{Majority Baseline}
Five  most frequent tags 
per domain from training, data are considered as Top1-Top5 predictions for the test data in order (Hit@1 to Hit@5). 
We introduce this baseline as the top few tags cover a large number of posts in each domain (Table \ref{tab:tag_post_coverage}).

\subsection{Feature-Based Models}
We use linear multi-label classifiers using the one-vs-all strategy with
\textit{tf-idf} 
and \textit{bag-of-word} features as  two baselines since most of the feature-based tag prediction models use either of them as features. We hypothesize that these models can leverage the high amount of tag-post overlap (Figure \ref{fig:tag_post_overlap}). Here we train the models for each domain with classes corresponding to all the unique tags.


\subsection{MetaTag Predictor Model (MP)} 
\begin{figure}
    \centering
    \includegraphics[width=0.95\columnwidth]{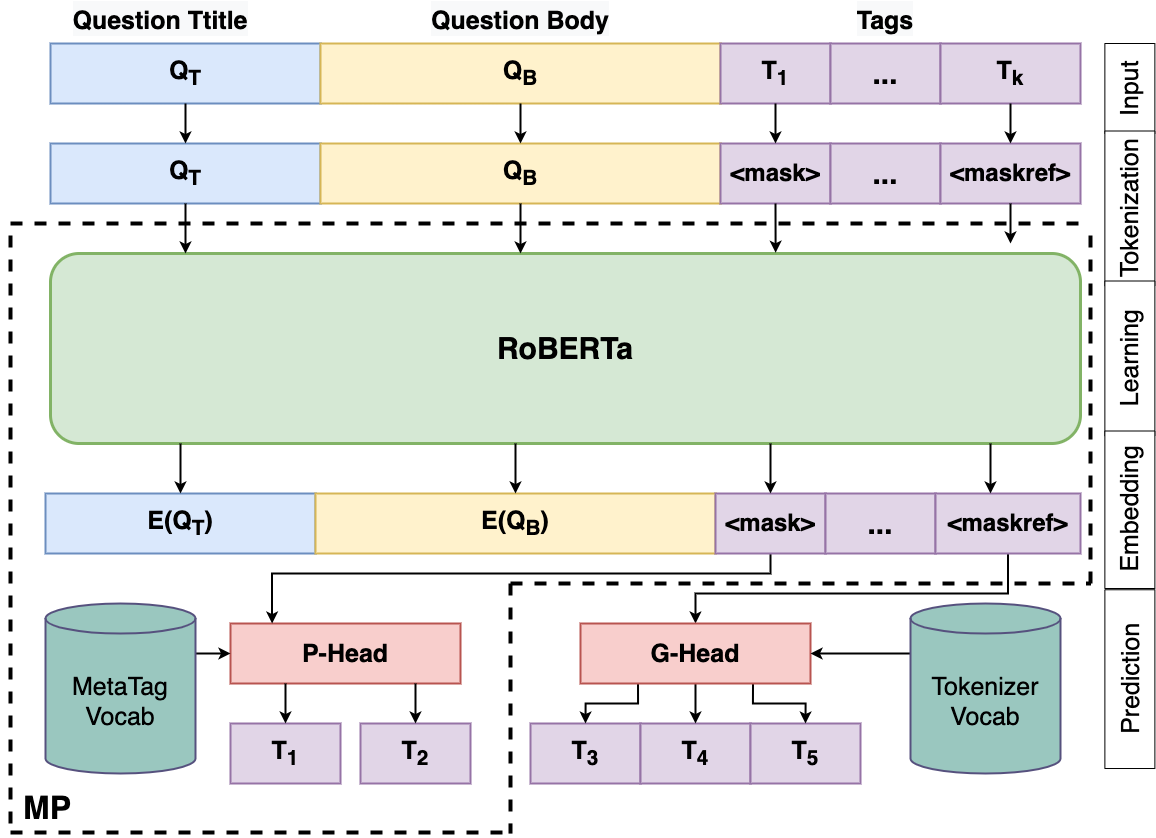}
    \caption{Meta-Refined Predictor Generator Model (MRPG), MetaTag Predictor (MP) model is highlighted in the dotted line having only the P-Head.
    }
    \label{fig:mrpg_model}
\end{figure}

In this model (Figure \ref{fig:mrpg_model}), we first select a vocabulary (\textit{MetaTag}) of tags based on a frequency analysis of the tag's post coverage per domain. Here, we consider popular tags as meta tags.
We formulate this multi-label classification task as a language model mask-filling task using \textit{pre-trained roberta-base} \cite{liu2019roberta} as the base of this model. We train separately for each domain.

\noindent
\textbf{Training:} 
We tokenize the question title ($Q_T$) and body ($Q_B$) and hide the tags from the MetaTag vocabulary with a mask token, \mask. These are concatenated and provided as input to the model.

\begin{center}
    $Q_T$ + $Q_B$ +\mask...\mask
\end{center}
\noindent
  This model is trained to predict those masks optimizing the prediction loss ($\mathcal{L}_{P}$) over all masked tokens ($\mathcal{L} = \mathcal{L}_{P} $). Here  the number of mask tokens may vary based on the post (shown as above). $\mathcal{L}$ is the total loss.

\noindent
\textbf{Inference:}
We tokenize  $Q_T$ and $Q_B$, and append five \mask \space tokens at the end, enforcing the model to predict exactly five tags for the post (the most probable tag for each position). This is because StackExchange allows a maximum of five tags to be associated with a question. This ensures that the model predicts the tags from the MetaTag vocabulary.

\begin{center}
$Q_T$ + $Q_B$ +\mask\mask\mask\mask\mask    
\end{center}

\subsection{Meta Refined Tag Predictor Generator Model (MRPG)}
This model (Figure \ref{fig:mrpg_model}) is similar to the MP model, with the additional ability to generate tags not present in the MetaTag vocabulary (OOV). In a more general sense, here the motivation is to develop a model capable of \textit{predicting} tags from a predefined set and \textit{generating} novel tags as well.

\noindent
\textbf{Training:}
Similar to MP model, we tokenize $Q_T$ and $Q_B$, and replace the tags present in the MetaTag vocabulary with \mask \space token. The rest of the tags (out-of-vocab or OOV) are tokenized and each token is replaced with a separate mask token,  \maskref. A \tagsep\space token is added to mark the boundaries (start and end) of these OOV tag tokens.
The model is trained on joint loss ($\mathcal{L}$) of meta tag prediction head loss ($\mathcal{L}_P$) and refined tag generation head loss ($\mathcal{L}_G$) given by $\mathcal{L} = \mathcal{L}_{P} + \mathcal{L}_{G}$.


\noindent
\textbf{Inference:} Our goal is to encourage the model to generate a combination of meta and refined tags. Based on our  tag-stability analysis (Section \ref{sec:tag_stability}), tag pair ordering analysis (Section \ref{sec:tag_pair_ordering}) and soft tag-hierarchy findings (Section \ref{sec:hierarchy}), we train the MRPG model to predict the first two tags from the MetaTag vocabulary and to generate the remaining three tags based on the user texts. We append two \mask\space tokens and a parameterized number of \maskref \space tokens with tokenized $Q_T$ and $Q_B$.

\begin{center}
$Q_T$ + $Q_B$ +\mask\mask\maskref\dots\maskref    
\end{center}

\noindent
\textbf{Tag Generation:} For each  \maskref\space tokens, MRPG generates one token from the tokenizer vocabulary following a  
greedy approach by selecting the most probable token. We concatenate the generated tokens between two \tagsep \space tokens and form a tag.  
We choose the most probable three generated refined tags based on our earlier data analysis and stack exchange tag limitations. However, for implementing this model to any other CQA platform, this number can be incremented or decremented based on the above-mentioned parameter. Also, there is no restriction in the model that will limit it to generating tags with more than 3 words. But they are rare for most of the domains, as can be seen from Table \ref{tab:tag_word_length}.
More details are in Appendix Section \ref{sec:app_decoding}.

\section{Experiments}

\subsection{Settings}
We split our dataset into train-dev-test in the ratio 70:10:20 based on a random seed value. In our experiments we build our model on top of the base version (125M parameters) of pre-trained roberta language model. We remove html tags (since these tags are irrelevant to \SE \space tags) from the user contents (question title and body) before tag prediction. We ran all experiments on 4 NVIDIA RTX A6000 GPUs (48GB GPU memory)  with a batch size of 60 and an input length of 256. We use AdamW \cite{loshchilov2017decoupled} optimizer, linear warmup scheduler, and a learning rate of 5e-5.

\subsection{Metrics:}
We define Hit@k (where $k=1,\dots,5$) as the percentage of posts where at least one predicted tags match with the actual tags for $k$ predictions. We generate at most 5 tag predictions in line with \SE's upper limit of tags. This metric aligns with our motivation of maximizing the probability that a user will be able to find at least one tag among the recommended fixed number of tags. Hence we do not consider other metrics like precision and recall.

\subsection{Performance Analysis}

\subsubsection{Baseline vs MP vs MRPG}

In Table \ref{tab:sota}, 
we compare our models with the baselines (mean of five different runs). The feature-based models, bag-of-word, and tf-idf models are able to achieve good performance for those domains where we found a high overlap between user  texts and tags. We find that our MP model shows improvements over the majority baseline and the feature-based models by a substantial margin (p-values $<$ 0.05 on Wilcoxon test) in Hit@5 performance. The MRPG model outperforms other methods in almost all the domains (significant improvements in 12 out of 17 domains). This is because it was able to generate tags outside the MetaTag vocabulary. In the \domain{biology} domain, the MP model performs better than MRPG. This might be because of the high tag reuse in this domain. 
All the model performance numbers (Hit@k for $k=1\dots5$) are present in Appendix Table \ref{tab:app_model_comp}. In this table, we observe that for Hit@1 MRPG model is always better than MP model.

\begin{table}[]
\caption{Performance Hit@5, 90\% Tag-Post Coverage. Significant improvements (p-values$<$0.05) of MRPG over MP are in bold. P-values are in Appendix \ref{sec:app_pvalues}}
\label{tab:sota}
\centering
\resizebox{0.95\columnwidth}{!}{%
\begin{tabular}{@{}l|rrr|rr@{}}
\toprule
\textbf{Domain} & \textbf{Majority} & \textbf{TF-IDF} & \textbf{Bag-of-Words} & \textbf{MP} & \textbf{MRPG} \\ \midrule
askubuntu &24.84 & 59.76$\pm$0.06 & 71.25$\pm$0.56 & 80.44$\pm$0.11 & \textbf{82.94}$\pm$0.15\\
aviation &35.05 & 55.12$\pm$0.29 & 65.58$\pm$0.64 & 77.09$\pm$0.44 & 77.63$\pm$0.56\\
biology &37.94 & 54.91$\pm$0.16 & 64.79$\pm$0.50 & 78.96$\pm$0.34 & 77.55$\pm$0.41\\
chemistry &48.89 & 58.76$\pm$0.17 & 68.09$\pm$0.46 & 77.66$\pm$0.10 & \textbf{79.17}$\pm$0.45\\
cooking &29.04 & 70.28$\pm$0.19 & 71.69$\pm$0.34 & 80.86$\pm$0.42 & \textbf{85.18}$\pm$0.29\\
electronics &20.68 & 57.80$\pm$0.11 & 70.12$\pm$0.13 & 77.51$\pm$0.26 & \textbf{81.30}$\pm$0.53\\
history &34.67 & 58.93$\pm$0.32 & 59.29$\pm$0.36 & 80.45$\pm$0.09 & 81.23$\pm$1.00\\
money &55.96 & 75.54$\pm$0.19 & 79.70$\pm$0.30 & 84.15$\pm$0.23 & \textbf{87.94}$\pm$0.42\\
movies &54.99 & 60.80$\pm$0.14 & 64.57$\pm$0.24 & 82.91$\pm$0.55 & 83.25$\pm$0.99\\
music &47.91 & 68.15$\pm$0.15 & 74.26$\pm$0.42 & 82.66$\pm$0.26 & \textbf{83.71}$\pm$0.51\\
philosophy &48.93 & 62.71$\pm$0.10 & 64.06$\pm$0.34 & 79.45$\pm$0.20 & 79.49$\pm$0.56\\
physics &39.98 & 66.81$\pm$0.16 & 79.59$\pm$0.17 & 81.12$\pm$0.22 & \textbf{86.34}$\pm$0.37\\
politics &64.16 & 81.50$\pm$0.21 & 83.37$\pm$0.73 & 86.29$\pm$0.25 & \textbf{90.98}$\pm$0.46\\
rpg &76.66 & 75.79$\pm$0.23 & 82.71$\pm$0.24 & 83.31$\pm$0.33 & \textbf{89.09}$\pm$0.16\\
scifi &62.24 & 80.48$\pm$0.10 & 85.88$\pm$0.21 & 85.91$\pm$0.11 & \textbf{91.53}$\pm$0.32\\
serverfault &29.84 & 62.83$\pm$0.06 & 73.07$\pm$0.20 & 81.66$\pm$0.16 & \textbf{85.82}$\pm$0.26\\
travel &48.31 & 76.82$\pm$0.48 & 83.73$\pm$0.27 & 83.96$\pm$0.12 & \textbf{89.50}$\pm$0.30\\
\bottomrule
\end{tabular}%
}
\end{table}

\subsubsection{Effects of Vocabulary Size Reduction}
\begin{figure}
    \centering
    \includegraphics[scale=0.33]{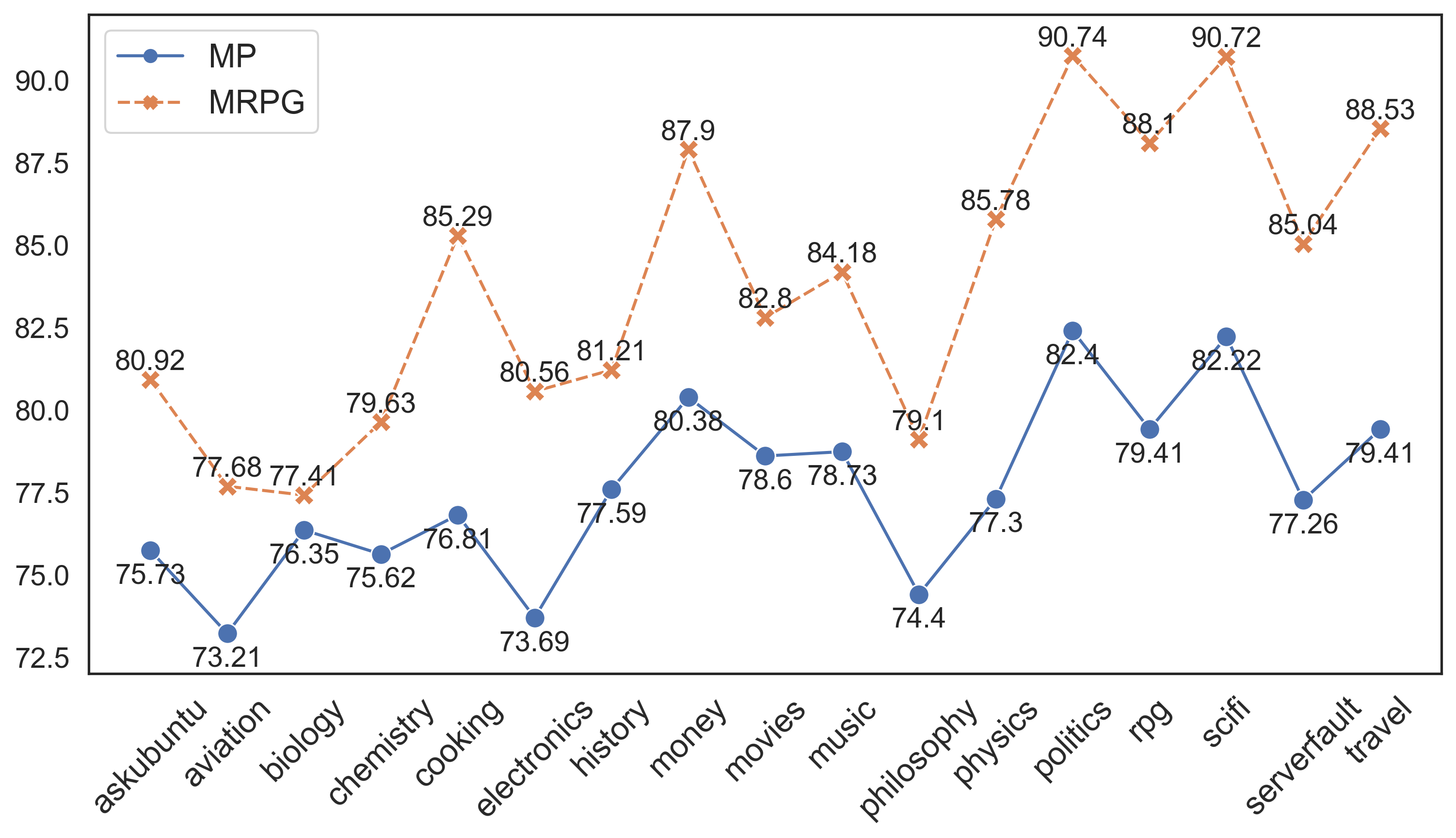}
    \caption{Effect of Vocab Size Reduction on MP vs MRPG for Hit@5 metric at 85\% post-coverage by tags}
    \label{fig:effect_vocabsize_models}
\end{figure}

We build the MetaTag vocab with 85\% post-coverage by tags ($\downarrow$5\%) and show the impact in Figure \ref{fig:effect_vocabsize_models}. 
We observe that \textit{the performance gap between MP and MRPG} at 90\% (Table \ref{tab:sota}) \textit{reduces} as vocab size decreases by 5\% (Figure \ref{fig:effect_vocabsize_models}) across all domains. 
This is because the MP model suffers the most (2-5\%) for this reduction. This is expected since MP's performance (by P-head) is based on how big the MetaTag vocabulary is. MRPG model, however, is \textit{robust to this vocabulary reduction}, i.e., the performance (Hit@5) only changes in the range 0-1.13\% with the exception of \domain{askubuntu} domain (2.26\%).
Details are in Appendix Table \ref{tab:app_vocab_diff_model_comp}.
Also with reduced vocab, the maximum performance difference is 9.12\% (travel) since it has more refined tags (Section \ref{sec:tag_stability}). The minimum difference is 1.06\% (\domain{biology}). Here the MRPG model could not take much advantage over MP because of high tag reusablity and fewer refined tags.

\begin{figure}
    \centering
    \includegraphics[scale=0.33]{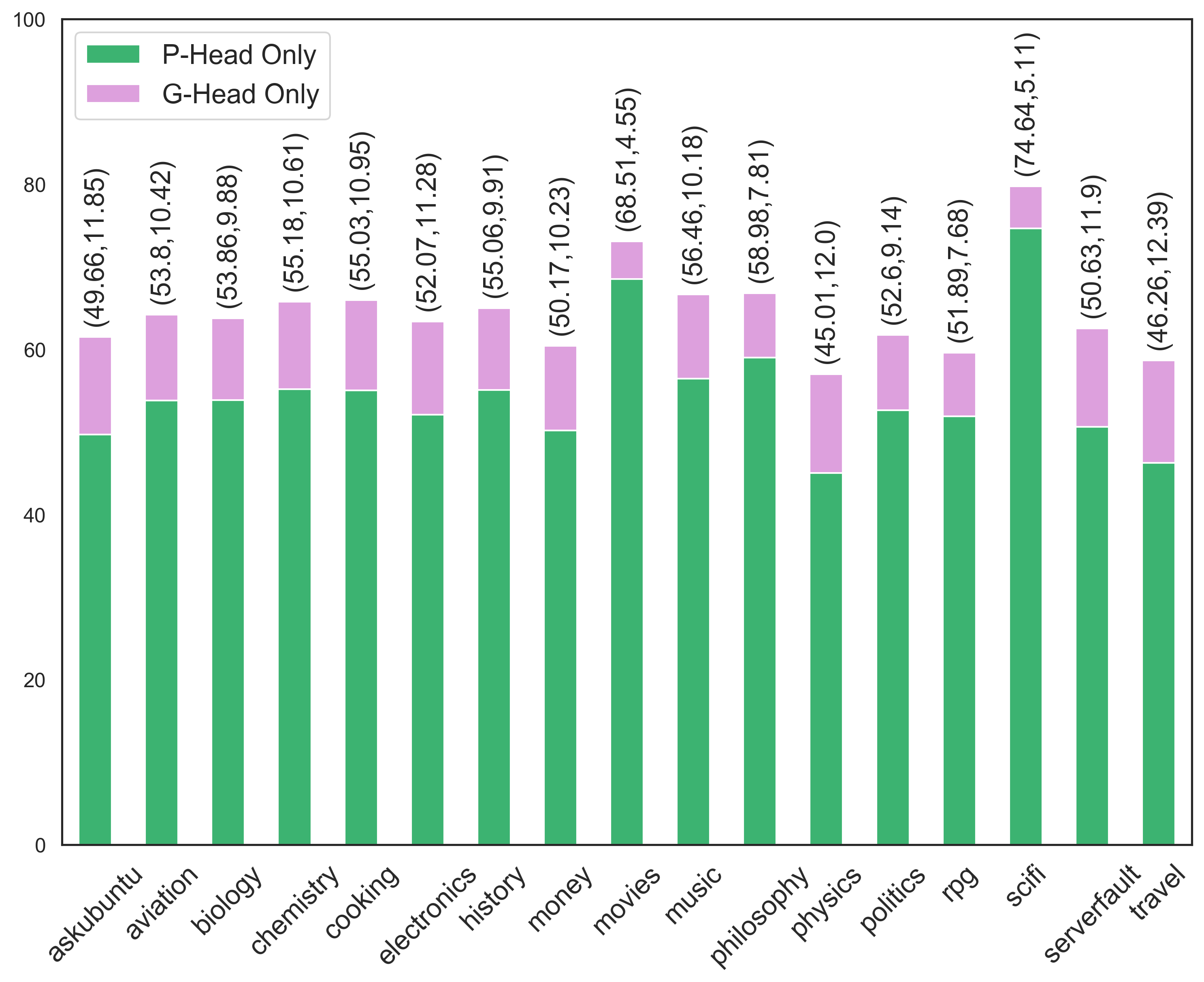}
    \caption{MRPG's Heads Contributions (Hit@5)}
    \label{fig:head_contrib}
\end{figure}

\subsubsection{Head Contribution of MRPG}
Figure \ref{fig:head_contrib} shows the contribution of P-Head and G-Head in the prediction performance (Hit@5 for 90\% coverage vocab). We extract for how many posts (\%) (1) only the P-Head correctly predicted at least one tag and (2) only the G-Head correctly predicted at least one tag.
P-Head's contributions were highest (45-74\%) since the MetaTag Vocabulary is created using popular tags in each domain. The G-Head was  able to predict at least one tag correctly for an extra 4-13\% of the posts. The effect of decreasing and increasing the MetaTag vocabulary size by 5\% change in tag-post coverage is shown in Appendix Table \ref{tab:head_contrib}. We observe that the G-Head's contribution increases up to 4\% (on vocab size decrease) and decreases up to 5\% (on vocab size increase). We also find that both the heads combined were able to suggest some non-overlapping tags in up to 33\% of the posts.

\subsubsection{Out-of-Vocabulary Tags Generation \%}
Table \ref{tab:oov_tag_stas} shows MRPG's performance in the prediction of tags outside  MetaTag Vocabulary for 90\% Tag-Post Coverage. \textit{\% Posts} shows the percentage of  posts where MRPG correctly predicted at least one OOV tag. It has the least contribution in two domains \domain{movies} (13.88\%) and \domain{scifi} (17.01\%). \textit{\% ALL Tags} and \textit{\% OOV Tags} shows that MRPG was able to correctly predict a considerable amount of OOV tags because of the generative head.

\begin{table*}[]
\caption{MRPG's Out-of-Vocab (OOV) Tag Prediction Match on Test. \textbf{\% Posts:} \% total posts where MRPG correctly predicted at least one OOV tag. \textbf{\% ALL Tags:} \% correctly predicted OOV tags out of total \#gold tags. \textbf{\% OOV Tags:} \% correctly predicted OOV tags out of total \#OOV gold tags.}
\label{tab:oov_tag_stas}
\centering
\resizebox{\textwidth}{!}{%
\begin{tabular}{@{}l|rrrrrrrrrrrrrrrrr@{}}
\toprule
Domains &
  \multicolumn{1}{c}{\rotatebox[origin=l]{90}{askubuntu}} &
  \multicolumn{1}{c}{\rotatebox[origin=l]{90}{aviation}} &
  \multicolumn{1}{c}{\rotatebox[origin=l]{90}{biology}} &
  \multicolumn{1}{c}{\rotatebox[origin=l]{90}{chemistry}} &
  \multicolumn{1}{c}{\rotatebox[origin=l]{90}{cooking}} &
  \multicolumn{1}{c}{\rotatebox[origin=l]{90}{electronics}} &
  \multicolumn{1}{c}{\rotatebox[origin=l]{90}{history}} &
  \multicolumn{1}{c}{\rotatebox[origin=l]{90}{money}} &
  \multicolumn{1}{c}{\rotatebox[origin=l]{90}{movies}} &
  \multicolumn{1}{c}{\rotatebox[origin=l]{90}{music}} &
  \multicolumn{1}{c}{\rotatebox[origin=l]{90}{philosophy}} &
  \multicolumn{1}{c}{\rotatebox[origin=l]{90}{physics}} &
  \multicolumn{1}{c}{\rotatebox[origin=l]{90}{politics}} &
  \multicolumn{1}{c}{\rotatebox[origin=l]{90}{rpg}} &
  \multicolumn{1}{c}{\rotatebox[origin=l]{90}{scifi}} &
  \multicolumn{1}{c}{\rotatebox[origin=l]{90}{serverfault}} &
  \multicolumn{1}{c}{\rotatebox[origin=l]{90}{travel}} \\ \midrule
\% Posts &31.38 & 	23.10 & 	22.09 & 	24.68 & 	29.21 & 	27.32 & 	22.37 & 	35.8 & 	13.88 & 	25.19 & 	19.10 & 	41.93 & 	34.35 & 	36.49 & 	17.01 & 	34.69 & 	43.59 
   \\
\% ALL Tags  &12.74 & 	9.49 & 	8.98 & 	11.31 & 	13.65 & 	10.73 & 	8.17 & 	12.74 & 	6.85 & 	10.60 & 	8.55 & 	15.32 & 	13.18 & 	13.92 & 	8.10 & 	13.62 & 	15.66 
   \\
\% OOV Tags  & 41.92 & 	28.03 & 	27.34 & 	37.84 & 	47.81 & 	31.39 & 	22.78 & 	33.21 & 	22.76 & 	36.87 & 	28.55 & 	36.55 & 	35.37 & 	43.04 & 	38.96 & 	36.38 & 	38.80 \\
   
   \bottomrule
   
\end{tabular}%
}
\end{table*}



\begin{figure}
    \centering
    \includegraphics[width=0.98\columnwidth]{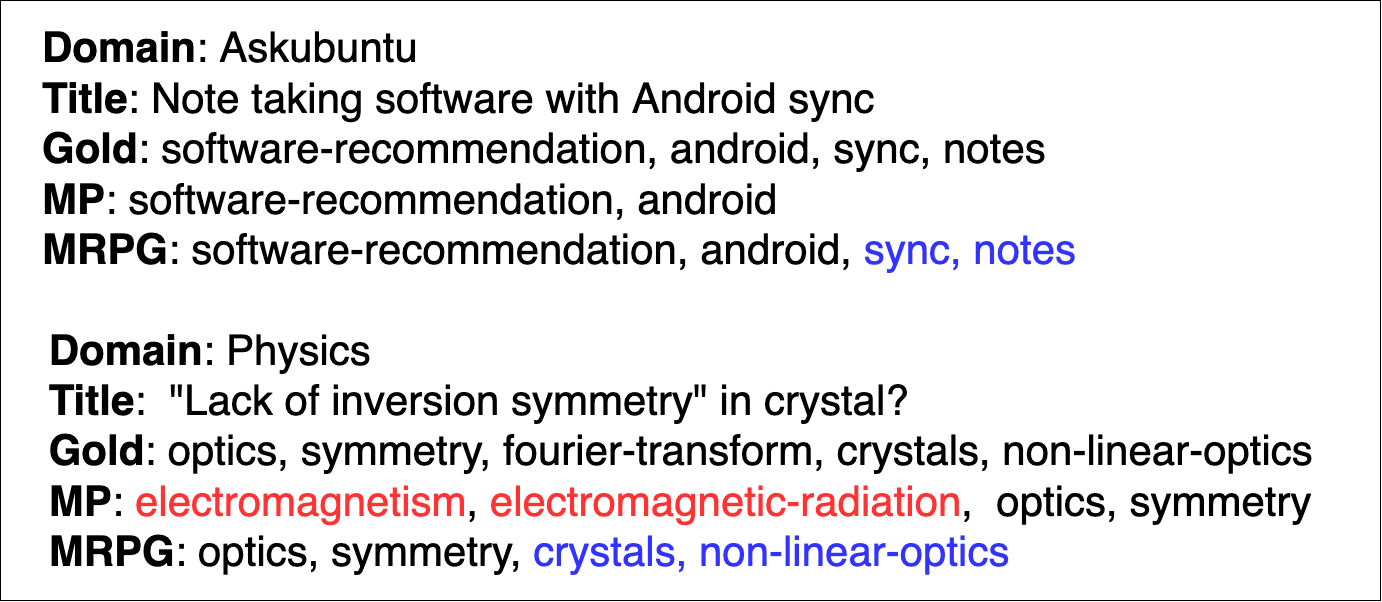}
    \caption{Case Studies}
    \label{fig:cqa_eg}
\end{figure}

\subsection{Case Studies}
We compare tag predictions of our methods in Figure \ref{fig:cqa_eg}. MRPG was able to generate two extra refined tags than MP in \domain{askubuntu} domain and was able to predict four out of five tags in \domain{physics} domain. Included below are examples for five other domains.

\label{sec:app_case_study}
\noindent
\begin{center}
\fbox{
    \small
    \begin{minipage}{0.9\columnwidth}
    \textbf{Domain: Physics} \\
    Title: Does matter become energy at the speed of light?\\
    Gold: special-relativity, speed-of-light, mass-energy, matter\\
    MP: special-relativity, energy, speed-of-light, mass\\
    MRPG: special-relativity, speed-of-light, mass-energy, matter
    \end{minipage}
}
\end{center}


\noindent
\begin{center}
\fbox{
    \small
    \begin{minipage}{0.9\columnwidth}
    \textbf{Domain: Travel} \\
    Title: Nigerian citizen (university student) was refused a UK visit visa due to lack of funds and connection to school - how to resolve?\\
    Gold: UK, visa-refusals, nigerian-citizens \\
    MP: visas, customs-and-immigration, visa-refusals, paperwork, standard-visitor-visas\\
    MRPG: uk, visa-refusals, nigerian-citizens
    \end{minipage}
}
\end{center}

\noindent
\begin{center}
\fbox{
    \small
    \begin{minipage}{0.9\columnwidth}
    \textbf{Domain: Music} \\
    Title: Piano tuning just under the absolute pitch\\
    Gold: piano, tuning \\
    MP: piano, tuning, maintenance\\
    MRPG: piano, tuning, alternative-tunings, pitch, relative-pitch
    \end{minipage}
}
\end{center}

\noindent
\begin{center}
\fbox{
    \small
    \begin{minipage}{0.9\columnwidth}
    \textbf{Domain: Biology} \\
    Title: Why aren't all infections immune-system resistant?\\
    Gold: evolution, microbiology, immunology, bacteriology \\
    MP: evolution, microbiology, bacteriology, bacteriology, immune-system\\
    MRPG: evolution, bacteriology, immunity, antibiotic-resistance
    \end{minipage}
}
\end{center}

\noindent
\begin{center}
\fbox{
    \small
    \begin{minipage}{0.9\columnwidth}
    \textbf{Domain: History} \\
    Title: Where to find a list of participants in The Crusades?\\
    Gold: middle-ages, crusades\\
    MP: middle-ages, middle-ages, europe, historiography\\
    MRPG: middle-ages, sources, crusades
    \end{minipage}
}
\end{center}

\subsection{Adaptability of the MP \& MRPG Architectures}
\label{sec:app_adaptability}
Both the MP and MRPG models can be adapted for use in other domains or in different public and private CQA platforms with specific tag-space restrictions. This can help in \textit{efficient question routing to area-experts for faster response time}, especially in private CQA platforms where the motivation of the community authority is to get queries resolved faster. Such adaptations can be done by customizing the MetaTag vocabulary based on prior behavioral analysis. Additionally, the \textit{number of meta and refined tags can be controlled} based on the domain and platform requirements without changes in architecture (through a parameter). Also, the MRPG model can be used in platforms where \textit{a soft-hierarchy of tags} is known, and routing requires the prediction of top-level tags and leaf tags. In such a scenario, the MetaTag vocabulary could be populated with only top-level tags, allowing the model to generate lower-level tags (from the tail of the tag distribution) based on user texts. With the combination of both types of tags, a query can be routed to a specific sub-area expert without overwhelming all the experts to a specific topic.

\section{Related Work}
\textbf{Community QA platform analysis: }
There have been several studies on Folksonomy \cite{vander2007folksonomy}, the practice of associating custom tags to questions in a social environment. 
Some of the prior works are:
a large-scale analysis of tags and their correlation with other tags \cite{fu2020modelling}, 
tag-distribution and tag-occurrence of 168 SE communities \cite{fu2020modelling}, quality analysis of SO \cite{singh2015stack}. User behavior analysis was done on Quora \cite{wang2013wisdom}, Yahoo Answers \cite{adamic2008knowledge}, Google Answers \cite{chen2010knowledge} and StackOverflow \cite{anderson2013steering}.
However, here we perform a large-scale study of tags, tag occurrences, and tag relation 
for 17 domains to understand how they have  some common properties in spite of being quite diverse, an observation similar to a prior work \cite{fu2020modelling}.

\noindent
\textbf{Community QA NLP Tasks:}
As the use of community QA platforms increased and with it the volume of community-created data, various NLP approaches were used to address some of the issues of each platform and also to understand behaviors of users. There have been various insights gathered through analysis of such communities. Similar Question Identification \cite{zhang2017detecting, zhang2018duplicate,vanam2021identifying,kumar2022transformer}, Similar Tag Identification \cite{beyer2015synonym,chen2019modeling}, Tag popularity prediction \cite{fu2017predicting}, Popular Question Prediction \cite{zhao2021hot}, Tag predictions \cite{lipczak2008tag, lipczak2010learning,wang2015tagcombine,wu2016tag2word,sonam2019tagstack,tang2019integral,wankerl2020f2tag,venktesh2021tagrec}, detecting anomalous tag combinations \cite{banerjee2019evaluating}, CQA entity linking \cite{li2022community}, expert recommendation \cite{tondulkar2018get,lv2021expert, menaha2021cluster, anandhan2022expert, krishna2022simplifying, askari2022expert, liu2022high}, question routing \cite{krishna2022topic}, identifying unclear questions \cite{trienes2019identifying}, automatic identification of best answers \cite{burel2012automatic} and tag-hierarchy predictions \cite{chen2019modeling} are some of the interesting tasks. 
We, perform a large-scale analysis with data over 10 years and across 17 diverse communities. We focus only on the tag-prediction NLP task for CQA platform.

\noindent
\textbf{Text Tagging: }
There are some feature-based machine learning approaches \cite{wang2015tagcombine, charte2015quinta, sonam2019tagstack, zangerle2011using, sigurbjornsson2008flickr, zangerle2011using,
lipczak2010learning, wu2016tag2word} and some deep learning approaches \cite{tang2019integral, li2020tagdc, wankerl2020f2tag} for tag prediction.  
Tagcombine \cite{wang2015tagcombine} uses 
software object similarity while TagStack \cite{sonam2019tagstack} uses tf-idf features with Naive Bayes classifier on StackOverflow texts.
QUINTA \cite{charte2015quinta} works on  6 \SE \space domains using KNN, \cite{zangerle2011using} on microblogging sites (Twitter) based on tweet-similarity, Tag2word \cite{wu2016tag2word} in math and StackOverflow domains using an LDA variant, \cite{lipczak2008tag, lipczak2010learning} on BibSonomy and StackOverflow datasets based on tag co-occurrence and user preference.  Among the deep learning methods,
F2Tag \cite{wankerl2020f2tag} is on math domains based on visual and textual formula representation, ITAG \cite{tang2019integral} is on the math domain using  RNN   and 
TagDC \cite{li2020tagdc} is based on software object similarity using an LSTM.
We here, predict a soft hierarchy of tags (predicting both meta and fine-grained tags) unlike the above-mentioned methods. 

\section{Conclusion}
We perform an in-depth analysis of 17  domains in a popular CQA platform, \SE, focusing on various aspects of question tagging such as domain diversity analysis, tag-space analysis, tag co-occurrence analysis, tag order, and tag positional stability. We present multiple insights into user behavior in assigning tags to the questions they post. Based on these findings we develop a tag prediction architecture that generates rarer and finer-grained tags in addition to popular tags from a pre-selected vocabulary. Our approach significantly out-perform feature-based baselines and also shows significant improvement in 12 domains when compared with vocabulary-based approach. 

\section{Limitations}
The analysis and its findings presented here are limited to 17 selected \SE\space domains considering their diversity. However, they may vary for the remaining 150 domains. Some of the findings (e.g. tag's positional stability) may vary for other CQA platforms which do not have any bounds on the number of tags. 
We use roberta-base and a smaller input size (256 tokens) for our experiments. With larger models and more context, the performance is expected to increase since more context usually leads to better learning by larger parameterized models.
We have ignored the answers in \SE\space for model training. We believe that indiscriminately selecting all answers as context for a question could be too noisy and if we were to select one or more appropriate answers, this would add complexity in choosing between the fastest answer, best answer, accepted answers, etc. We consider this as a separate area of research and future work.
We randomly sampled the data for each domain to create the train and test split to show that our MRPG model is capable of both predicting and generating tags. Splitting with respect to timestamp would require tag temporal analysis and tag-evolution which we consider as a future area of research.


\section{Ethical Statement}
This work analyzes various aspects of aggregate tagging behavior of users on a popular community question-answering platform \SE. The data is publicly provided by \SE\space as an anonymized dump of all user-contributed content on the Stack Exchange network. The data is cc-by-sa 4.0 licensed, and intended to be shared and remixed. 
No specific user has been identified and no user-level information (user name etc.) has been used for this work. We only used the \textit{Post.xml} extracted from the \SE\space dumps and do not use any user profile statistics. The aggregate user behavior has been analyzed with respect to tagging and user-generated questions. Based on these findings a tag predictor model has been developed. 
The data has not been modified or redistributed as part of this research.

\bibliographystyle{ACM-Reference-Format}
\bibliography{sample-base}


\begin{thebibliography}{43}


\ifx \showCODEN    \undefined \def \showCODEN     #1{\unskip}     \fi
\ifx \showDOI      \undefined \def \showDOI       #1{#1}\fi
\ifx \showISBNx    \undefined \def \showISBNx     #1{\unskip}     \fi
\ifx \showISBNxiii \undefined \def \showISBNxiii  #1{\unskip}     \fi
\ifx \showISSN     \undefined \def \showISSN      #1{\unskip}     \fi
\ifx \showLCCN     \undefined \def \showLCCN      #1{\unskip}     \fi
\ifx \shownote     \undefined \def \shownote      #1{#1}          \fi
\ifx \showarticletitle \undefined \def \showarticletitle #1{#1}   \fi
\ifx \showURL      \undefined \def \showURL       {\relax}        \fi
\providecommand\bibfield[2]{#2}
\providecommand\bibinfo[2]{#2}
\providecommand\natexlab[1]{#1}
\providecommand\showeprint[2][]{arXiv:#2}

\bibitem[Adamic et~al\mbox{.}(2008)]%
        {adamic2008knowledge}
\bibfield{author}{\bibinfo{person}{Lada~A Adamic}, \bibinfo{person}{Jun Zhang},
  \bibinfo{person}{Eytan Bakshy}, {and} \bibinfo{person}{Mark~S Ackerman}.}
  \bibinfo{year}{2008}\natexlab{}.
\newblock \showarticletitle{Knowledge sharing and yahoo answers: everyone knows
  something}. In \bibinfo{booktitle}{\emph{Proceedings of the 17th
  international conference on World Wide Web}}. \bibinfo{pages}{665--674}.
\newblock


\bibitem[Anandhan et~al\mbox{.}(2022)]%
        {anandhan2022expert}
\bibfield{author}{\bibinfo{person}{Anitha Anandhan},
  \bibinfo{person}{Maizatul~Akmar Ismail}, {and} \bibinfo{person}{Liyana
  Shuib}.} \bibinfo{year}{2022}\natexlab{}.
\newblock \showarticletitle{EXPERT RECOMMENDATION THROUGH TAG RELATIONSHIP IN
  COMMUNITY QUESTION ANSWERING}.
\newblock \bibinfo{journal}{\emph{Malaysian Journal of Computer Science}}
  \bibinfo{volume}{35}, \bibinfo{number}{3} (\bibinfo{year}{2022}),
  \bibinfo{pages}{201--221}.
\newblock


\bibitem[Anderson et~al\mbox{.}(2013)]%
        {anderson2013steering}
\bibfield{author}{\bibinfo{person}{Ashton Anderson}, \bibinfo{person}{Daniel
  Huttenlocher}, \bibinfo{person}{Jon Kleinberg}, {and} \bibinfo{person}{Jure
  Leskovec}.} \bibinfo{year}{2013}\natexlab{}.
\newblock \showarticletitle{Steering user behavior with badges}. In
  \bibinfo{booktitle}{\emph{Proceedings of the 22nd international conference on
  World Wide Web}}. \bibinfo{pages}{95--106}.
\newblock


\bibitem[Askari et~al\mbox{.}(2022)]%
        {askari2022expert}
\bibfield{author}{\bibinfo{person}{Arian Askari}, \bibinfo{person}{Suzan
  Verberne}, {and} \bibinfo{person}{Gabriella Pasi}.}
  \bibinfo{year}{2022}\natexlab{}.
\newblock \showarticletitle{Expert Finding in Legal Community Question
  Answering}. In \bibinfo{booktitle}{\emph{European Conference on Information
  Retrieval}}. Springer, \bibinfo{pages}{22--30}.
\newblock


\bibitem[Banerjee et~al\mbox{.}(2019)]%
        {banerjee2019evaluating}
\bibfield{author}{\bibinfo{person}{Rohan Banerjee}, \bibinfo{person}{Sailaja
  Rajanala}, {and} \bibinfo{person}{Manish Singh}.}
  \bibinfo{year}{2019}\natexlab{}.
\newblock \showarticletitle{Evaluating the Choice of Tags in CQA Sites}. In
  \bibinfo{booktitle}{\emph{International Conference on Database Systems for
  Advanced Applications}}. Springer, \bibinfo{pages}{625--640}.
\newblock


\bibitem[Beyer and Pinzger(2015)]%
        {beyer2015synonym}
\bibfield{author}{\bibinfo{person}{Stefanie Beyer} {and}
  \bibinfo{person}{Martin Pinzger}.} \bibinfo{year}{2015}\natexlab{}.
\newblock \showarticletitle{Synonym suggestion for tags on stack overflow}. In
  \bibinfo{booktitle}{\emph{2015 IEEE 23rd International Conference on Program
  Comprehension}}. IEEE, \bibinfo{pages}{94--103}.
\newblock


\bibitem[Burel et~al\mbox{.}(2012)]%
        {burel2012automatic}
\bibfield{author}{\bibinfo{person}{Gr{\'e}goire Burel}, \bibinfo{person}{Yulan
  He}, {and} \bibinfo{person}{Harith Alani}.} \bibinfo{year}{2012}\natexlab{}.
\newblock \showarticletitle{Automatic identification of best answers in online
  enquiry communities}. In \bibinfo{booktitle}{\emph{Extended Semantic Web
  Conference}}. Springer, \bibinfo{pages}{514--529}.
\newblock


\bibitem[Charte et~al\mbox{.}(2015)]%
        {charte2015quinta}
\bibfield{author}{\bibinfo{person}{Francisco Charte},
  \bibinfo{person}{Antonio~J Rivera}, \bibinfo{person}{Mar{\'\i}a~J del Jesus},
  {and} \bibinfo{person}{Francisco Herrera}.} \bibinfo{year}{2015}\natexlab{}.
\newblock \showarticletitle{QUINTA: A question tagging assistant to improve the
  answering ratio in electronic forums}. In \bibinfo{booktitle}{\emph{Ieee
  eurocon 2015-international conference on computer as a tool (eurocon)}}.
  IEEE, \bibinfo{pages}{1--6}.
\newblock


\bibitem[Chen et~al\mbox{.}(2019)]%
        {chen2019modeling}
\bibfield{author}{\bibinfo{person}{Hui Chen}, \bibinfo{person}{John Coogle},
  {and} \bibinfo{person}{Kostadin Damevski}.} \bibinfo{year}{2019}\natexlab{}.
\newblock \showarticletitle{Modeling stack overflow tags and topics as a
  hierarchy of concepts}.
\newblock \bibinfo{journal}{\emph{Journal of Systems and Software}}
  \bibinfo{volume}{156} (\bibinfo{year}{2019}), \bibinfo{pages}{283--299}.
\newblock


\bibitem[Chen et~al\mbox{.}(2010)]%
        {chen2010knowledge}
\bibfield{author}{\bibinfo{person}{Yan Chen}, \bibinfo{person}{Teck-Hua Ho},
  {and} \bibinfo{person}{Yong-mi Kim}.} \bibinfo{year}{2010}\natexlab{}.
\newblock \showarticletitle{Knowledge market design: A field experiment at
  Google Answers}.
\newblock \bibinfo{journal}{\emph{Journal of Public Economic Theory}}
  \bibinfo{volume}{12}, \bibinfo{number}{4} (\bibinfo{year}{2010}),
  \bibinfo{pages}{641--664}.
\newblock


\bibitem[Fu et~al\mbox{.}(2017)]%
        {fu2017predicting}
\bibfield{author}{\bibinfo{person}{Chenbo Fu}, \bibinfo{person}{Yongli Zheng},
  \bibinfo{person}{Shidi Li}, \bibinfo{person}{Qi Xuan}, {and}
  \bibinfo{person}{Zhongyuan Ruan}.} \bibinfo{year}{2017}\natexlab{}.
\newblock \showarticletitle{Predicting the popularity of tags in StackExchange
  QA communities}. In \bibinfo{booktitle}{\emph{2017 International Workshop on
  Complex Systems and Networks (IWCSN)}}. IEEE, \bibinfo{pages}{90--95}.
\newblock


\bibitem[Fu et~al\mbox{.}(2020)]%
        {fu2020modelling}
\bibfield{author}{\bibinfo{person}{Xiang Fu}, \bibinfo{person}{Shangdi Yu},
  {and} \bibinfo{person}{Austin~R Benson}.} \bibinfo{year}{2020}\natexlab{}.
\newblock \showarticletitle{Modelling and analysis of tagging networks in Stack
  Exchange communities}.
\newblock \bibinfo{journal}{\emph{Journal of Complex Networks}}
  \bibinfo{volume}{8}, \bibinfo{number}{5} (\bibinfo{year}{2020}),
  \bibinfo{pages}{cnz045}.
\newblock


\bibitem[Hollander et~al\mbox{.}(2013)]%
        {hollander2013nonparametric}
\bibfield{author}{\bibinfo{person}{Myles Hollander}, \bibinfo{person}{Douglas~A
  Wolfe}, {and} \bibinfo{person}{Eric Chicken}.}
  \bibinfo{year}{2013}\natexlab{}.
\newblock \bibinfo{booktitle}{\emph{Nonparametric statistical methods}}.
\newblock \bibinfo{publisher}{John Wiley \& Sons}.
\newblock


\bibitem[Krishna and Antulov-Fantulin(2022)]%
        {krishna2022simplifying}
\bibfield{author}{\bibinfo{person}{Vaibhav Krishna} {and} \bibinfo{person}{Nino
  Antulov-Fantulin}.} \bibinfo{year}{2022}\natexlab{}.
\newblock \showarticletitle{Simplifying Sparse Expert Recommendation by
  Revisiting Graph Diffusion}.
\newblock \bibinfo{journal}{\emph{arXiv preprint arXiv:2208.02438}}
  (\bibinfo{year}{2022}).
\newblock


\bibitem[Krishna et~al\mbox{.}(2022)]%
        {krishna2022topic}
\bibfield{author}{\bibinfo{person}{Vaibhav Krishna}, \bibinfo{person}{Vaiva
  Vasiliauskaite}, {and} \bibinfo{person}{Nino Antulov-Fantulin}.}
  \bibinfo{year}{2022}\natexlab{}.
\newblock \showarticletitle{Topic Community Based Temporal Expertise for
  Question Routing}.
\newblock \bibinfo{journal}{\emph{arXiv preprint arXiv:2207.01753}}
  (\bibinfo{year}{2022}).
\newblock


\bibitem[Kumar and Chauhan(2022)]%
        {kumar2022transformer}
\bibfield{author}{\bibinfo{person}{Shobhan Kumar} {and} \bibinfo{person}{Arun
  Chauhan}.} \bibinfo{year}{2022}\natexlab{}.
\newblock \showarticletitle{A Transformer Based Encodings for Detection of
  Semantically Equivalent Questions in cQA}.
\newblock \bibinfo{journal}{\emph{Comput. J.}} (\bibinfo{year}{2022}).
\newblock


\bibitem[Li et~al\mbox{.}(2020)]%
        {li2020tagdc}
\bibfield{author}{\bibinfo{person}{Can Li}, \bibinfo{person}{Ling Xu},
  \bibinfo{person}{Meng Yan}, {and} \bibinfo{person}{Yan Lei}.}
  \bibinfo{year}{2020}\natexlab{}.
\newblock \showarticletitle{TagDC: A tag recommendation method for software
  information sites with a combination of deep learning and collaborative
  filtering}.
\newblock \bibinfo{journal}{\emph{Journal of Systems and Software}}
  \bibinfo{volume}{170} (\bibinfo{year}{2020}), \bibinfo{pages}{110783}.
\newblock


\bibitem[Li et~al\mbox{.}(2022)]%
        {li2022community}
\bibfield{author}{\bibinfo{person}{Yuhan Li}, \bibinfo{person}{Wei Shen},
  \bibinfo{person}{Jianbo Gao}, {and} \bibinfo{person}{Yadong Wang}.}
  \bibinfo{year}{2022}\natexlab{}.
\newblock \showarticletitle{Community Question Answering Entity Linking via
  Leveraging Auxiliary Data}.
\newblock \bibinfo{journal}{\emph{arXiv preprint arXiv:2205.11917}}
  (\bibinfo{year}{2022}).
\newblock


\bibitem[Lipczak(2008)]%
        {lipczak2008tag}
\bibfield{author}{\bibinfo{person}{Marek Lipczak}.}
  \bibinfo{year}{2008}\natexlab{}.
\newblock \showarticletitle{Tag recommendation for folksonomies oriented
  towards individual users}.
\newblock \bibinfo{journal}{\emph{ECML PKDD discovery challenge}}
  \bibinfo{volume}{84} (\bibinfo{year}{2008}), \bibinfo{pages}{2008}.
\newblock


\bibitem[Lipczak and Milios(2010)]%
        {lipczak2010learning}
\bibfield{author}{\bibinfo{person}{Marek Lipczak} {and}
  \bibinfo{person}{Evangelos Milios}.} \bibinfo{year}{2010}\natexlab{}.
\newblock \showarticletitle{Learning in efficient tag recommendation}. In
  \bibinfo{booktitle}{\emph{Proceedings of the fourth ACM conference on
  Recommender systems}}. \bibinfo{pages}{167--174}.
\newblock


\bibitem[Liu et~al\mbox{.}(2019)]%
        {liu2019roberta}
\bibfield{author}{\bibinfo{person}{Yinhan Liu}, \bibinfo{person}{Myle Ott},
  \bibinfo{person}{Naman Goyal}, \bibinfo{person}{Jingfei Du},
  \bibinfo{person}{Mandar Joshi}, \bibinfo{person}{Danqi Chen},
  \bibinfo{person}{Omer Levy}, \bibinfo{person}{Mike Lewis},
  \bibinfo{person}{Luke Zettlemoyer}, {and} \bibinfo{person}{Veselin
  Stoyanov}.} \bibinfo{year}{2019}\natexlab{}.
\newblock \showarticletitle{Roberta: A robustly optimized bert pretraining
  approach}.
\newblock \bibinfo{journal}{\emph{arXiv preprint arXiv:1907.11692}}
  (\bibinfo{year}{2019}).
\newblock


\bibitem[Liu et~al\mbox{.}(2022)]%
        {liu2022high}
\bibfield{author}{\bibinfo{person}{Yue Liu}, \bibinfo{person}{Weize Tang},
  \bibinfo{person}{Zitu Liu}, \bibinfo{person}{Lin Ding}, {and}
  \bibinfo{person}{Aihua Tang}.} \bibinfo{year}{2022}\natexlab{}.
\newblock \showarticletitle{High-quality domain expert finding method in CQA
  based on multi-granularity semantic analysis and interest drift}.
\newblock \bibinfo{journal}{\emph{Information Sciences}}  \bibinfo{volume}{596}
  (\bibinfo{year}{2022}), \bibinfo{pages}{395--413}.
\newblock


\bibitem[Loshchilov and Hutter(2017)]%
        {loshchilov2017decoupled}
\bibfield{author}{\bibinfo{person}{Ilya Loshchilov} {and}
  \bibinfo{person}{Frank Hutter}.} \bibinfo{year}{2017}\natexlab{}.
\newblock \showarticletitle{Decoupled weight decay regularization}.
\newblock \bibinfo{journal}{\emph{arXiv preprint arXiv:1711.05101}}
  (\bibinfo{year}{2017}).
\newblock


\bibitem[Lv et~al\mbox{.}(2021)]%
        {lv2021expert}
\bibfield{author}{\bibinfo{person}{Xiaoqi Lv}, \bibinfo{person}{Ke Ji},
  \bibinfo{person}{Zhenxiang Chen}, \bibinfo{person}{Kun Ma},
  \bibinfo{person}{Jun Wu}, \bibinfo{person}{Yidong Li}, {and}
  \bibinfo{person}{Guandong Xu}.} \bibinfo{year}{2021}\natexlab{}.
\newblock \showarticletitle{Expert Recommendations with Temporal Dynamics of
  User Interest in CQA}. In \bibinfo{booktitle}{\emph{International Conference
  on Web Information Systems Engineering}}. Springer,
  \bibinfo{pages}{645--652}.
\newblock


\bibitem[Menaha et~al\mbox{.}(2021)]%
        {menaha2021cluster}
\bibfield{author}{\bibinfo{person}{R Menaha}, \bibinfo{person}{VE Jayanthi},
  \bibinfo{person}{N Krishnaraj}, {et~al\mbox{.}}}
  \bibinfo{year}{2021}\natexlab{}.
\newblock \showarticletitle{A Cluster-based Approach for Finding Domain wise
  Experts in Community Question Answering System}. In
  \bibinfo{booktitle}{\emph{Journal of Physics: Conference Series}},
  Vol.~\bibinfo{volume}{1767}. IOP Publishing, \bibinfo{pages}{012035}.
\newblock


\bibitem[Parnell et~al\mbox{.}(2011)]%
        {parnell2011biostar}
\bibfield{author}{\bibinfo{person}{Laurence~D Parnell}, \bibinfo{person}{Pierre
  Lindenbaum}, \bibinfo{person}{Khader Shameer},
  \bibinfo{person}{Giovanni~Marco Dall'Olio}, \bibinfo{person}{Daniel~C Swan},
  \bibinfo{person}{Lars~Juhl Jensen}, \bibinfo{person}{Simon~J Cockell},
  \bibinfo{person}{Brent~S Pedersen}, \bibinfo{person}{Mary~E Mangan},
  \bibinfo{person}{Christopher~A Miller}, {et~al\mbox{.}}}
  \bibinfo{year}{2011}\natexlab{}.
\newblock \showarticletitle{BioStar: an online question \& answer resource for
  the bioinformatics community}.
\newblock \bibinfo{journal}{\emph{PLoS computational biology}}
  \bibinfo{volume}{7}, \bibinfo{number}{10} (\bibinfo{year}{2011}),
  \bibinfo{pages}{e1002216}.
\newblock


\bibitem[Sigurbj{\"o}rnsson and Van~Zwol(2008)]%
        {sigurbjornsson2008flickr}
\bibfield{author}{\bibinfo{person}{B{\"o}rkur Sigurbj{\"o}rnsson} {and}
  \bibinfo{person}{Roelof Van~Zwol}.} \bibinfo{year}{2008}\natexlab{}.
\newblock \showarticletitle{Flickr tag recommendation based on collective
  knowledge}. In \bibinfo{booktitle}{\emph{Proceedings of the 17th
  international conference on World Wide Web}}. \bibinfo{pages}{327--336}.
\newblock


\bibitem[Singh et~al\mbox{.}(2015)]%
        {singh2015stack}
\bibfield{author}{\bibinfo{person}{Sanjay Singh} {et~al\mbox{.}}}
  \bibinfo{year}{2015}\natexlab{}.
\newblock \showarticletitle{Is Stack Overflow Overflowing With Questions and
  Tags}.
\newblock \bibinfo{journal}{\emph{arXiv preprint arXiv:1508.03601}}
  (\bibinfo{year}{2015}).
\newblock


\bibitem[Sonam et~al\mbox{.}(2019)]%
        {sonam2019tagstack}
\bibfield{author}{\bibinfo{person}{Sonam Sonam}, \bibinfo{person}{Ayushi
  Verma}, \bibinfo{person}{Sangeeta Lal}, {and} \bibinfo{person}{Neetu
  Sardana}.} \bibinfo{year}{2019}\natexlab{}.
\newblock \showarticletitle{TagStack: Automated system for predicting tags in
  stackoverflow}. In \bibinfo{booktitle}{\emph{2019 International Conference on
  Signal Processing and Communication (ICSC)}}. IEEE,
  \bibinfo{pages}{223--228}.
\newblock


\bibitem[Tang et~al\mbox{.}(2019)]%
        {tang2019integral}
\bibfield{author}{\bibinfo{person}{Shijie Tang}, \bibinfo{person}{Yuan Yao},
  \bibinfo{person}{Suwei Zhang}, \bibinfo{person}{Feng Xu},
  \bibinfo{person}{Tianxiao Gu}, \bibinfo{person}{Hanghang Tong},
  \bibinfo{person}{Xiaohui Yan}, {and} \bibinfo{person}{Jian Lu}.}
  \bibinfo{year}{2019}\natexlab{}.
\newblock \showarticletitle{An integral tag recommendation model for textual
  content}. In \bibinfo{booktitle}{\emph{Proceedings of the AAAI Conference on
  Artificial Intelligence}}, Vol.~\bibinfo{volume}{33}.
  \bibinfo{pages}{5109--5116}.
\newblock


\bibitem[Tondulkar et~al\mbox{.}(2018)]%
        {tondulkar2018get}
\bibfield{author}{\bibinfo{person}{Rohan Tondulkar}, \bibinfo{person}{Manisha
  Dubey}, {and} \bibinfo{person}{Maunendra~Sankar Desarkar}.}
  \bibinfo{year}{2018}\natexlab{}.
\newblock \showarticletitle{Get me the best: predicting best answerers in
  community question answering sites}. In \bibinfo{booktitle}{\emph{Proceedings
  of the 12th ACM Conference on Recommender Systems}}.
  \bibinfo{pages}{251--259}.
\newblock


\bibitem[Trienes and Balog(2019)]%
        {trienes2019identifying}
\bibfield{author}{\bibinfo{person}{Jan Trienes} {and}
  \bibinfo{person}{Krisztian Balog}.} \bibinfo{year}{2019}\natexlab{}.
\newblock \showarticletitle{Identifying unclear questions in community question
  answering websites}. In \bibinfo{booktitle}{\emph{European conference on
  information retrieval}}. Springer, \bibinfo{pages}{276--289}.
\newblock


\bibitem[Vanam and Pulipati(2021)]%
        {vanam2021identifying}
\bibfield{author}{\bibinfo{person}{Divya Vanam} {and}
  \bibinfo{person}{Venkateswara~Rao Pulipati}.}
  \bibinfo{year}{2021}\natexlab{}.
\newblock \showarticletitle{Identifying Duplicate Questions in Community
  Question Answering Forums Using Machine Learning Approaches}.
\newblock In \bibinfo{booktitle}{\emph{Machine Learning Technologies and
  Applications}}. \bibinfo{publisher}{Springer}, \bibinfo{pages}{131--140}.
\newblock


\bibitem[Vander~Wal(2007)]%
        {vander2007folksonomy}
\bibfield{author}{\bibinfo{person}{Thomas Vander~Wal}.}
  \bibinfo{year}{2007}\natexlab{}.
\newblock \bibinfo{title}{Folksonomy}.
\newblock
\newblock


\bibitem[Venktesh et~al\mbox{.}(2021)]%
        {venktesh2021tagrec}
\bibfield{author}{\bibinfo{person}{V Venktesh}, \bibinfo{person}{Mukesh
  Mohania}, {and} \bibinfo{person}{Vikram Goyal}.}
  \bibinfo{year}{2021}\natexlab{}.
\newblock \showarticletitle{TagRec: Automated Tagging of Questions with
  Hierarchical Learning Taxonomy}. In \bibinfo{booktitle}{\emph{Joint European
  Conference on Machine Learning and Knowledge Discovery in Databases}}.
  Springer, \bibinfo{pages}{381--396}.
\newblock


\bibitem[Wang et~al\mbox{.}(2013)]%
        {wang2013wisdom}
\bibfield{author}{\bibinfo{person}{Gang Wang}, \bibinfo{person}{Konark Gill},
  \bibinfo{person}{Manish Mohanlal}, \bibinfo{person}{Haitao Zheng}, {and}
  \bibinfo{person}{Ben~Y Zhao}.} \bibinfo{year}{2013}\natexlab{}.
\newblock \showarticletitle{Wisdom in the social crowd: an analysis of quora}.
  In \bibinfo{booktitle}{\emph{Proceedings of the 22nd international conference
  on World Wide Web}}. \bibinfo{pages}{1341--1352}.
\newblock


\bibitem[Wang et~al\mbox{.}(2015)]%
        {wang2015tagcombine}
\bibfield{author}{\bibinfo{person}{Xin-Yu Wang}, \bibinfo{person}{Xin Xia},
  {and} \bibinfo{person}{David Lo}.} \bibinfo{year}{2015}\natexlab{}.
\newblock \showarticletitle{Tagcombine: Recommending tags to contents in
  software information sites}.
\newblock \bibinfo{journal}{\emph{Journal of Computer Science and Technology}}
  \bibinfo{volume}{30}, \bibinfo{number}{5} (\bibinfo{year}{2015}),
  \bibinfo{pages}{1017--1035}.
\newblock


\bibitem[Wankerl et~al\mbox{.}(2020)]%
        {wankerl2020f2tag}
\bibfield{author}{\bibinfo{person}{Sebastian Wankerl}, \bibinfo{person}{Gerhard
  G{\"o}tz}, {and} \bibinfo{person}{Andreas Hotho}.}
  \bibinfo{year}{2020}\natexlab{}.
\newblock \showarticletitle{f2tag—Can Tags be Predicted Using Formulas?}. In
  \bibinfo{booktitle}{\emph{2020 19th IEEE International Conference on Machine
  Learning and Applications (ICMLA)}}. IEEE, \bibinfo{pages}{565--571}.
\newblock


\bibitem[Wu et~al\mbox{.}(2016)]%
        {wu2016tag2word}
\bibfield{author}{\bibinfo{person}{Yong Wu}, \bibinfo{person}{Yuan Yao},
  \bibinfo{person}{Feng Xu}, \bibinfo{person}{Hanghang Tong}, {and}
  \bibinfo{person}{Jian Lu}.} \bibinfo{year}{2016}\natexlab{}.
\newblock \showarticletitle{Tag2word: Using tags to generate words for content
  based tag recommendation}. In \bibinfo{booktitle}{\emph{Proceedings of the
  25th ACM international on conference on information and knowledge
  management}}. \bibinfo{pages}{2287--2292}.
\newblock


\bibitem[Zangerle et~al\mbox{.}(2011)]%
        {zangerle2011using}
\bibfield{author}{\bibinfo{person}{Eva Zangerle}, \bibinfo{person}{Wolfgang
  Gassler}, {and} \bibinfo{person}{G{\"u}nther Specht}.}
  \bibinfo{year}{2011}\natexlab{}.
\newblock \showarticletitle{Using tag recommendations to homogenize
  folksonomies in microblogging environments}. In
  \bibinfo{booktitle}{\emph{International conference on social informatics}}.
  Springer, \bibinfo{pages}{113--126}.
\newblock


\bibitem[Zhang et~al\mbox{.}(2017)]%
        {zhang2017detecting}
\bibfield{author}{\bibinfo{person}{Wei~Emma Zhang}, \bibinfo{person}{Quan~Z
  Sheng}, \bibinfo{person}{Jey~Han Lau}, {and} \bibinfo{person}{Ermyas Abebe}.}
  \bibinfo{year}{2017}\natexlab{}.
\newblock \showarticletitle{Detecting duplicate posts in programming QA
  communities via latent semantics and association rules}. In
  \bibinfo{booktitle}{\emph{Proceedings of the 26th International Conference on
  World Wide Web}}. \bibinfo{pages}{1221--1229}.
\newblock


\bibitem[Zhang et~al\mbox{.}(2018)]%
        {zhang2018duplicate}
\bibfield{author}{\bibinfo{person}{Wei~Emma Zhang}, \bibinfo{person}{Quan~Z
  Sheng}, \bibinfo{person}{Jey~Han Lau}, \bibinfo{person}{Ermyas Abebe}, {and}
  \bibinfo{person}{Wenjie Ruan}.} \bibinfo{year}{2018}\natexlab{}.
\newblock \showarticletitle{Duplicate detection in programming question
  answering communities}.
\newblock \bibinfo{journal}{\emph{ACM Transactions on Internet Technology
  (TOIT)}} \bibinfo{volume}{18}, \bibinfo{number}{3} (\bibinfo{year}{2018}),
  \bibinfo{pages}{1--21}.
\newblock


\bibitem[Zhao et~al\mbox{.}(2021)]%
        {zhao2021hot}
\bibfield{author}{\bibinfo{person}{Li~Xian Zhao}, \bibinfo{person}{Li Zhang},
  {and} \bibinfo{person}{Jing Jiang}.} \bibinfo{year}{2021}\natexlab{}.
\newblock \showarticletitle{Hot question prediction in Stack Overflow}.
\newblock \bibinfo{journal}{\emph{IET Software}} \bibinfo{volume}{15},
  \bibinfo{number}{1} (\bibinfo{year}{2021}), \bibinfo{pages}{90--106}.
\newblock


\end{thebibliography}

\appendix


\section{Domain Statistics}
\label{sec:app_domain_stats}
Table \ref{app:tab_more_stats} shows more details about domain diversity apart from those mentioned in the main section \ref{sec:comm_div}. We can see \textit{cooking} and \textit{rpg} are the domains with the least number of questions with no answers ($<$5\%) which indicates the experts in these domains are very active. The science domains have more than 15\% questions with no answers which shows that special knowledge is required to answer such questions. $maxview$ and $maxans$ show the maximum limit of users who viewed the questions and the maximum number of answers that a question has. \textit{no accept ans} shows the percentage of posts that have not been accepted by the askers as answers. This gives an indication of whether askers are active and also whether the answers are satisfactory.

\begin{table}[]
\caption{Tag Statistics: AvgTLen - Average Tag Length}
\label{tab:app_tag_stats}
\resizebox{\linewidth}{!}{%
\begin{tabular}{l|lr|lr|r}
\hline
\multicolumn{1}{l}{\multirow{2}{*}{Domains}} &
  \multicolumn{2}{l}{Longest Tag (number of characters)} &
  \multicolumn{2}{l}{Shortest Tag} &
  \multicolumn{1}{l}{\multirow{2}{*}{AvgTLen}}  \\ \cline{2-5}
\multicolumn{1}{l}{} &
  \multicolumn{1}{l}{Tag} &
  \multicolumn{1}{l}{Size} &
  \multicolumn{1}{l}{Tag} &
  \multicolumn{1}{l}{Size} &
  \multicolumn{1}{l}{} \\ \hline
 askubuntu & windows-subsystem-for-linux & 27 & c & 1 & 8.17 \\ 
aviation & performance-based-navigation & 28 & cg & 2 & 10.53 \\ 
biology & neurodegenerative-disorders & 27 & ph & 2 & 10.97 \\ 
chemistry & differential-scanning-calorimetry & 33 & ph & 2 & 12.75 \\ 
cooking & please-remove-this-tag & 22 & ue & 2 & 8.56 \\ 
electronics & semiconductor-process-technology & 32 & c & 1 & 8.80 \\ 
history & articles-of-confederation & 25 & art & 3 & 9.83 \\ 
money & health-reimbursement-arrangement & 32 & w9 & 2 & 10.87 \\ 
movies & valerian-city-of-a-thousand-planets & 35 & m & 1 & 13.66 \\ 
music & solid-body-electric-guitars & 27 & dj & 2 & 9.62 \\ 
philosophy & philosophy-of-political-science & 31 & art & 3 & 11.17 \\ 
physics & heisenberg-uncertainty-principle & 32 & air & 3 & 13.39 \\ 
politics & immigration-customs-enforcement & 31 & alp & 3 & 11.12 \\ 
rpg & werewolf-the-apocalypse-2nd-edition & 35 & e6 & 2 & 12.29 \\ 
scifi & the-hitchhikers-guide-to-the-galaxy & 35 & dc & 2 & 13.31 \\ 
serverfault & google-cloud-internal-load-balancer & 35 & 3g & 2 & 8.87 \\
travel & new-zealand-permanent-resident & 30 & eu & 2 & 9.39 \\ 
\bottomrule
\end{tabular}%
}
\end{table}

\section{Tag Length Analysis} 
Table \ref{tab:app_tag_stats} shows the maximum and minimum length tags in each domain. We also see that the average tag length of the \domain{movies} and \domain{physics} domain are the highest. We find that often the movie names or physics topics are longer than three words leading to an increase in average tag length.

\begin{table*}[]
\caption{Domain Statistics}
\label{app:tab_more_stats}
\resizebox{0.9\textwidth}{!}{%
\begin{tabular}{@{}l|r|r|r|r|r|r|r|r|r|r|r@{}}
\toprule
Domain & Q & T & Q/T & AVGT & NOANS (\%) & NOSCORES (\%) & NO ACCEPT ANS (\%) & MAXANS & MAXVIEW & VIEWGT100 & \#ASKERS \\ \midrule askubuntu & 371800 & 3121 & 119.13 & 2.78 & 23.47 & 37.21 & 66.99 & 82 & 5409384 & 1093 & 201912 \\
aviation & 20345 & 1002 & 20.3 & 2.56 & 7.02 & 9.82 & 46.66 & 18 & 219002 & 12 & 7066 \\
biology & 25671 & 739 & 34.74 & 2.58 & 20.73 & 15.16 & 56.13 & 11 & 445257 & 11 & 12089 \\
chemistry & 37476 & 375 & 99.94 & 2.37 & 19.36 & 16.86 & 59.12 & 11 & 1077991 & 7 & 17202 \\
cooking & 24513 & 833 & 29.43 & 2.3 & 4.6 & 11.79 & 50.57 & 85 & 1619295 & 13 & 12413 \\
electronics & 152980 & 2226 & 68.72 & 2.77 & 9.13 & 40.87 & 50.94 & 38 & 591616 & 36 & 61869 \\
history & 12562 & 813 & 15.45 & 2.84 & 9.85 & 4.6 & 49.94 & 34 & 994376 & 19 & 5296 \\
money & 32648 & 995 & 32.81 & 3.11 & 7.91 & 17.73 & 54.56 & 25 & 821144 & 37 & 18010 \\
movies & 20749 & 4348 & 4.77 & 2.09 & 9.48 & 2.96 & 39.08 & 19 & 1183407 & 30 & 6931 \\
music & 20925 & 512 & 40.87 & 2.52 & 3.24 & 10.96 & 49.62 & 25 & 611990 & 5 & 10447 \\
philosophy & 15624 & 559 & 27.95 & 2.4 & 11 & 15.27 & 63.73 & 31 & 250018 & 6 & 6640 \\
physics & 180166 & 893 & 201.75 & 3.17 & 17.49 & 29.54 & 57.08 & 49 & 847876 & 131 & 59774 \\
politics & 12416 & 739 & 16.8 & 2.9 & 6.81 & 5.55 & 48.72 & 27 & 833812 & 27 & 3970 \\
rpg & 42693 & 1195 & 35.73 & 2.91 & 4.41 & 2.26 & 32.39 & 44 & 865197 & 56 & 11541 \\
scifi & 62987 & 3433 & 18.35 & 2.25 & 10.62 & 2.45 & 42.87 & 34 & 1430390 & 153 & 22717 \\
serverfault & 299895 & 3814 & 78.63 & 2.9 & 11.68 & 37.05 & 51.93 & 160 & 2478923 & 327 & 130214 \\
travel & 42201 & 1891 & 22.32 & 3.28 & 11.2 & 8.19 & 59.48 & 30 & 430504 & 42 & 24895 \\
\bottomrule
\end{tabular}%
}
\end{table*}

\begin{figure*}
    \centering
    \includegraphics[width=\textwidth]{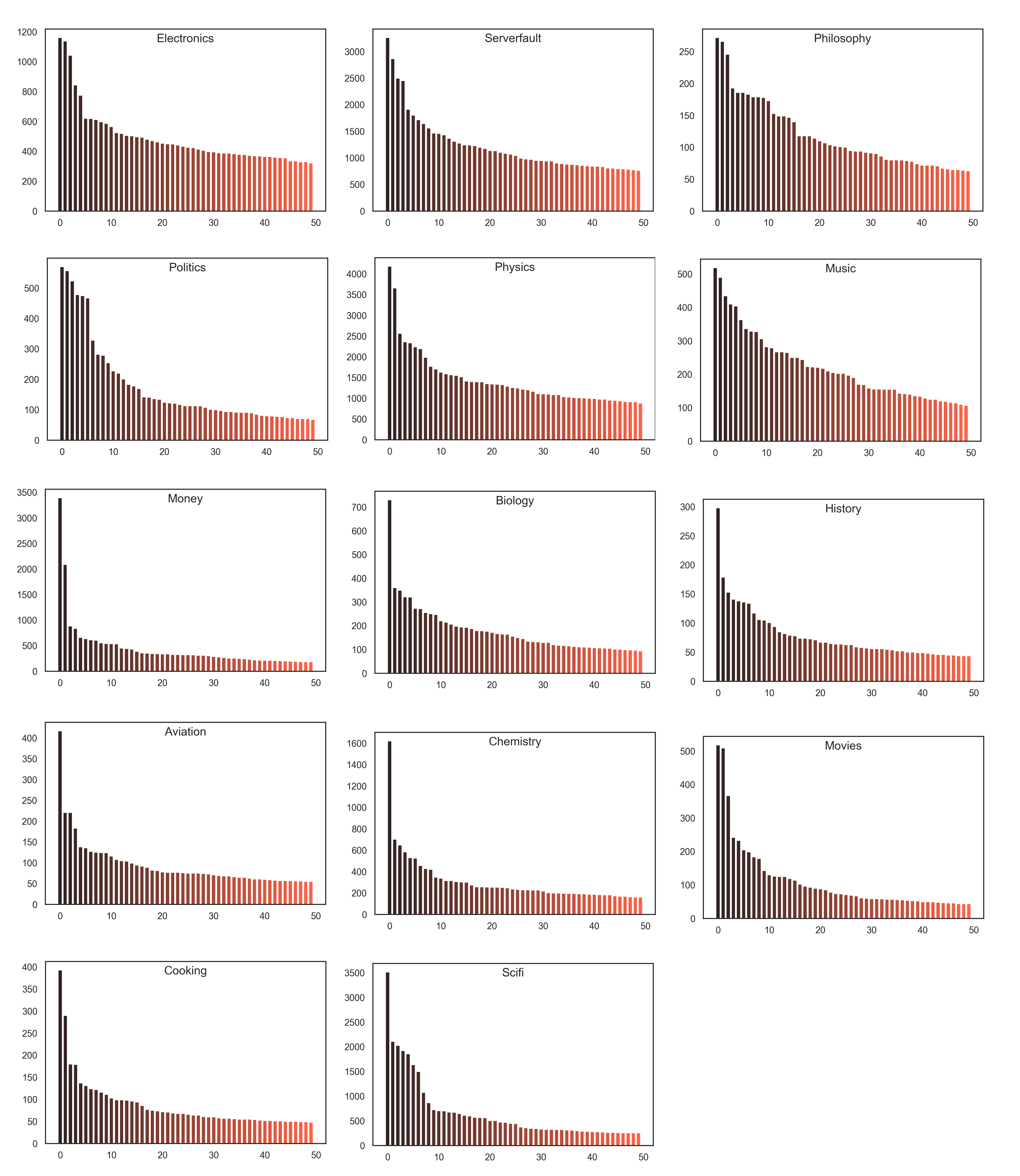}
    \caption{Tag Co-Occurrence Distributions (Rest),Y-Axis: \# posts the tag pairs appear in, X-Axis: Top-50 Tags}
    \label{fig:app_tag_cooccur}
\end{figure*}


\section{Tag Co-Occurrence Distribution Analysis}
\label{sec:tag_cooccur}
\begin{figure*}
    \centering
    \includegraphics[width=\textwidth, height=2.3cm]{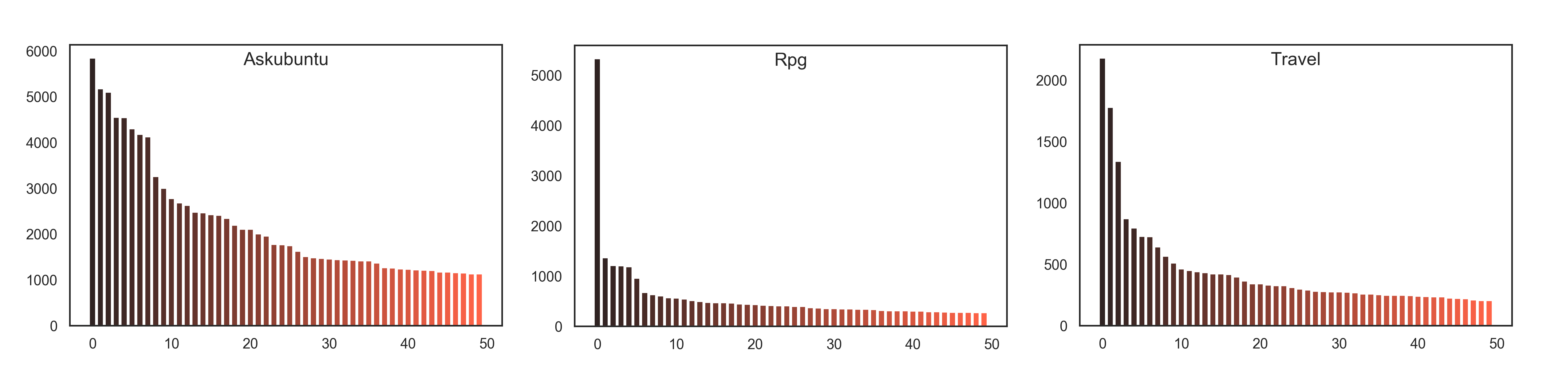}
    \caption{Tag Co-Occurrence Distribution Patterns,Y-Axis: \# posts the tag pairs appear in, X-Axis: Top-50 Tags}
    \label{fig:tag_cooccur_3}
\end{figure*}
We analyzed the distribution of the top-50 frequently occurring tag pairs in each domain (Figure \ref{fig:app_tag_cooccur}, \ref{fig:tag_cooccur_3}). We observe three main patterns: (1) \textit{Smooth Distribution} (2) \textit{Spike in Top-1} (3) \textit{Spikes in top few pairs}.
Larger domains like \domain{askubuntu}, \domain{serverfault}, \domain{electronics}, and \domain{physics}, have smooth distributions.
Some of the smaller domains like \domain{politics}, \domain{philosophy}, and \domain{music} also show this behavior, which we believe is because, in these domains, the questions have  
fine-grained topics. In domains like rpg, money, history, aviation, biology, chemistry, the tags of the most frequent tag pair that appears in abundance are generic in nature. 
Finally, in domains like movies, scifi, cooking and travel, few tag pairs dominate the distributions, indicating their popularity in such smaller domains.

\section{Tag Co-Occurrence Examples}
\label{sec:app_tag_cooccur_top5}
Table \ref{tab:app_tag_cooccur_top5} shows the most frequent tag pairs that appear in each domain.

\begin{table*}[]

\caption{Top-5 Most Frequent Tag Pairs}
\label{tab:app_tag_cooccur_top5}
\centering

\resizebox{0.9\textwidth}{!}{%
\begin{tabular}{@{}l|l@{}}
\toprule
    Domain & Top-5 Most Frequent Tag Pairs \\
    \midrule
    askubuntu & (boot, grub2), (boot, dual-boot), (dual-boot, grub2), (bash, command-line), (apt, package-management) \\
    aviation & (aerodynamics, aircraft-design), (aircraft-design, wing), (aerodynamics, wing), (aircraft-design, aircraft-performance),\\
    & (air-traffic-control, faa-regulations) \\
    biology & (entomology, species-identification), (species-identification, zoology), (botany, species-identification), \\ 
    & (neurophysiology, neuroscience), (biochemistry, molecular-biology) \\
    chemistry & (organic-chemistry, reaction-mechanism), (physical-chemistry, thermodynamics), (aromatic-compounds, organic-chemistry), \\
    & (nomenclature, organic-chemistry), (carbonyl-compounds, organic-chemistry) \\
    cooking & (baking, bread), (baking, cake), (baking, cookies), (baking, substitutions), (bread, dough) \\
    electronics & (current, voltage), (pcb, pcb-design), (power, power-supply), (batteries, battery-charging), (microcontroller, pic) \\
    history & (nazi-germany, world-war-two), (united-states, world-war-two), (europe, middle-ages), \\
    & (japan, world-war-two), (military, world-war-two) \\
    money & (taxes, united-states), (income-tax, united-states), (401k, united-states), (income-tax, taxes), (tax-deduction, united-states) \\
    movies & (character, plot-explanation), (marvel-cinematic-universe, plot-explanation), \\
    & (game-of-thrones, plot-explanation), (analysis, plot-explanation), (avengers-infinity-war, marvel-cinematic-universe) \\
    music & (chords, theory), (chord-theory, chords), (harmony, theory), (scales, theory), (chord-theory, theory) \\
    philosophy & (logic, philosophy-of-mathematics), (epistemology, philosophy-of-science), (fallacies, logic), \\
    & (logic, symbolic-logic), (metaphysics, ontology) \\
    physics & (homework-and-exercises, newtonian-mechanics), (forces, newtonian-mechanics), (hilbert-space, quantum-mechanics), \\ 
    & (operators, quantum-mechanics), (quantum-mechanics, wavefunction) \\
    politics & (donald-trump, united-states), (president, united-states), (presidential-election, united-states), \\
    & (congress, united-states), (election, united-states) \\
    rpg & (dnd-5e, spells), (dnd-5e, magic-items), (class-feature, dnd-5e), (dnd-5e, monsters), (pathfinder-1e, spells) \\
    scifi & (short-stories, story-identification), (marvel, marvel-cinematic-universe), (books, story-identification), \\
    & (the-lord-of-the-rings, tolkiens-legendarium), (novel, story-identification) \\
    serverfault & (linux, ubuntu), (centos, linux), (amazon-ec2, amazon-web-services), (linux, networking), (apache-2.2, php) \\
    travel & (uk, visas), (schengen, visas), (usa, visas), (customs-and-immigration, usa), (indian-citizens, visas) \\
 \bottomrule
\end{tabular}%
}
\end{table*}

\section{Tag Distributions}
\begin{figure*}
    \centering
    \includegraphics[width=0.85\textwidth]{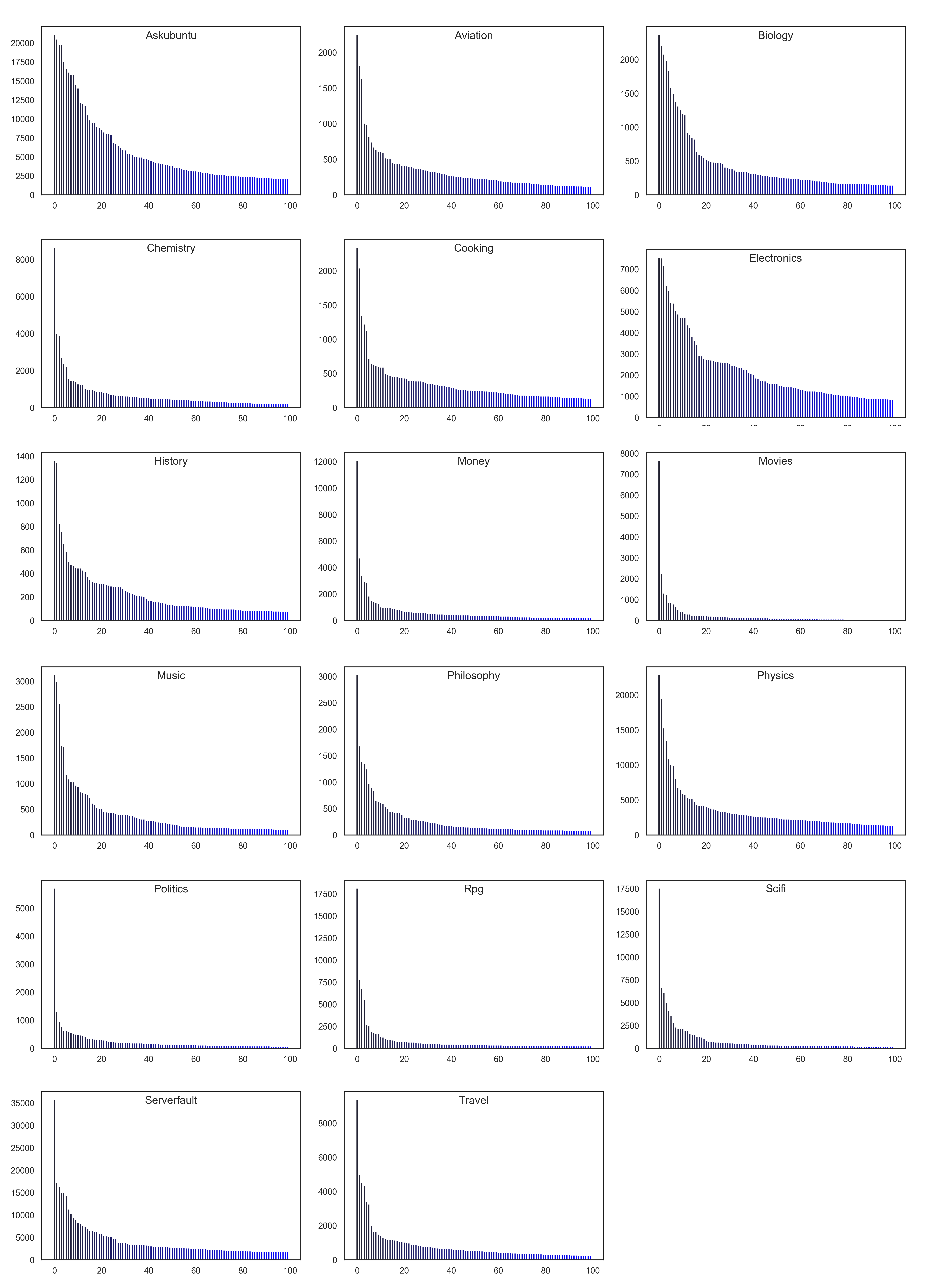}
    \caption{Tag Distributions, Y-Axis: \# posts the tag appear in, X-Axis: Top-100 Tags}
    \label{fig:app_tag_distrib}
\end{figure*}

Figure \ref{fig:app_tag_distrib} shows the distribution of top-100 most frequent tags in each domain.

\section{Tag Ordering Example:}
\label{sec:app_tag_ordering}

Tables \ref{tab:app_tagorder_1}, \ref{tab:app_tagorder_2}, and \ref{tab:app_tagorder_3} show top-10 most frequently occurring tag pairs in each domain. On analyzing manually, we  found that in most of the cases meta-tag appears before the refined tags.
\begin{table*}[]
\caption{Tag Ordering Statistics - First 6 domains}
\label{tab:app_tagorder_1}
\small
\resizebox{0.75\textwidth}{!}{%
\begin{tabular}{@{}lrlrlr@{}}
\toprule
Domain & Total & Order-1 & \% & Order-2 & \% \\ \midrule

askubuntu  & 5845 & (boot,grub2) & 99.93 & (grub2,boot) & 0.07 \\
askubuntu & 5174 & (boot,dual-boot) & 99.96 & (dual-boot,boot) & 0.04 \\
askubuntu & 5104 & (dual-boot,grub2) & 91.12 & (grub2,dual-boot) & 8.88 \\
askubuntu & 4552 & (bash,command-line) & 1.89 & (command-line,bash) & 98.11 \\
askubuntu & 4547 & (apt,package-management) & 98.53 & (package-management,apt) & 1.47 \\
askubuntu & 4304 & (networking,wireless) & 70.07 & (wireless,networking) & 29.93 \\
askubuntu & 4178 & (dual-boot,partitioning) & 97.75 & (partitioning,dual-boot) & 2.25 \\
askubuntu & 4128 & (drivers,nvidia) & 99.93 & (nvidia,drivers) & 0.07 \\
askubuntu & 3257 & (networking,server) & 97.97 & (server,networking) & 2.03 \\
askubuntu & 3003 & (bash,scripts) & 99.9 & (scripts,bash) & 0.1 \\ \hline
 
aviation  & 417 & (aerodynamics,aircraft-design) & 1.68 & (aircraft-design,aerodynamics) & 98.32 \\
aviation & 221 & (aircraft-design,wing) & 100 & (wing,aircraft-design) & 0 \\
aviation & 221 & (aerodynamics,wing) & 100 & (wing,aerodynamics) & 0 \\
aviation & 183 & (aircraft-design,aircraft-performance) & 100 & (aircraft-performance,aircraft-design) & 0 \\
aviation & 138 & (air-traffic-control,faa-regulations) & 0 & (faa-regulations,air-traffic-control) & 100 \\
aviation & 136 & (faa-regulations,instrument-flight-rules) & 100 & (instrument-flight-rules,faa-regulations) & 0 \\
aviation & 127 & (aerodynamics,lift) & 100 & (lift,aerodynamics) & 0 \\
aviation & 125 & (aerodynamics,airfoil) & 100 & (airfoil,aerodynamics) & 0 \\
aviation & 124 & (air-traffic-control,radio-communications) & 100 & (radio-communications,air-traffic-control) & 0 \\
aviation & 124 & (aerodynamics,aircraft-performance) & 100 & (aircraft-performance,aerodynamics) & 0 \\
 \hline
 
biology  & 731 & (entomology,species-identification) & 10.81 & (species-identification,entomology) & 89.19 \\
biology & 361 & (species-identification,zoology) & 76.18 & (zoology,species-identification) & 23.82 \\
biology & 350 & (botany,species-identification) & 44.29 & (species-identification,botany) & 55.71 \\
biology & 322 & (neurophysiology,neuroscience) & 0 & (neuroscience,neurophysiology) & 100 \\
biology & 321 & (biochemistry,molecular-biology) & 99.69 & (molecular-biology,biochemistry) & 0.31 \\
biology & 274 & (dna,genetics) & 4.38 & (genetics,dna) & 95.62 \\
biology & 272 & (evolution,genetics) & 37.5 & (genetics,evolution) & 62.5 \\
biology & 256 & (botany,plant-physiology) & 98.05 & (plant-physiology,botany) & 1.95 \\
biology & 251 & (entomology,zoology) & 0.8 & (zoology,entomology) & 99.2 \\
biology & 247 & (cell-biology,molecular-biology) & 1.21 & (molecular-biology,cell-biology) & 98.79 \\
 \hline
 
 chemistry & 1621 & (organic-chemistry,reaction-mechanism) & 100 & (reaction-mechanism,organic-chemistry) & 0 \\
chemistry & 703 & (physical-chemistry,thermodynamics) & 99.43 & (thermodynamics,physical-chemistry) & 0.57 \\
chemistry & 648 & (aromatic-compounds,organic-chemistry) & 0 & (organic-chemistry,aromatic-compounds) & 100 \\
chemistry & 585 & (nomenclature,organic-chemistry) & 0 & (organic-chemistry,nomenclature) & 100 \\
chemistry & 529 & (carbonyl-compounds,organic-chemistry) & 0 & (organic-chemistry,carbonyl-compounds) & 100 \\
chemistry & 526 & (acid-base,organic-chemistry) & 0 & (organic-chemistry,acid-base) & 100 \\
chemistry & 457 & (organic-chemistry,synthesis) & 100 & (synthesis,organic-chemistry) & 0 \\
chemistry & 429 & (organic-chemistry,stereochemistry) & 100 & (stereochemistry,organic-chemistry) & 0 \\
chemistry & 420 & (acid-base,ph) & 100 & (ph,acid-base) & 0 \\
chemistry & 348 & (acid-base,inorganic-chemistry) & 0 & (inorganic-chemistry,acid-base) & 100 \\
 \hline
 
cooking  & 393 & (baking,bread) & 99.75 & (bread,baking) & 0.25 \\
cooking & 290 & (baking,cake) & 100 & (cake,baking) & 0 \\
cooking & 180 & (baking,cookies) & 100 & (cookies,baking) & 0 \\
cooking & 179 & (baking,substitutions) & 91.06 & (substitutions,baking) & 8.94 \\
cooking & 137 & (bread,dough) & 91.24 & (dough,bread) & 8.76 \\
cooking & 131 & (bread,sourdough) & 100 & (sourdough,bread) & 0 \\
cooking & 124 & (baking,dough) & 100 & (dough,baking) & 0 \\
cooking & 122 & (bread,yeast) & 96.72 & (yeast,bread) & 3.28 \\
cooking & 116 & (baking,oven) & 100 & (oven,baking) & 0 \\
cooking & 111 & (dough,pizza) & 91.89 & (pizza,dough) & 8.11 \\
 \hline
 electronics & 1161 & (current,voltage) & 0.6 & (voltage,current) & 99.4 \\
electronics & 1138 & (pcb,pcb-design) & 100 & (pcb-design,pcb) & 0 \\
electronics & 1043 & (power,power-supply) & 0.48 & (power-supply,power) & 99.52 \\
electronics & 844 & (batteries,battery-charging) & 100 & (battery-charging,batteries) & 0 \\
electronics & 775 & (microcontroller,pic) & 98.58 & (pic,microcontroller) & 1.42 \\
electronics & 620 & (amplifier,operational-amplifier) & 3.87 & (operational-amplifier,amplifier) & 96.13 \\
electronics & 619 & (power-supply,switch-mode-power-supply) & 100 & (switch-mode-power-supply,power-supply) & 0 \\
electronics & 612 & (bjt,transistors) & 0.49 & (transistors,bjt) & 99.51 \\
electronics & 598 & (mosfet,transistors) & 0.17 & (transistors,mosfet) & 99.83 \\
electronics & 587 & (arduino,microcontroller) & 86.03 & (microcontroller,arduino) & 13.97 \\
\bottomrule
\end{tabular}%
}
\end{table*}

\begin{table*}[]
\caption{Tag Ordering Statistics - Second 6 domains}
\label{tab:app_tagorder_2}
\small
\resizebox{0.85\textwidth}{!}{%
\begin{tabular}{@{}lrlrlr@{}}
\toprule
Domain & Total & Order-1 & \% & Order-2 & \% \\ \midrule
history & 298 & (nazi-germany,world-war-two) & 0.67 & (world-war-two,nazi-germany) & 99.33 \\
history & 179 & (united-states,world-war-two) & 94.41 & (world-war-two,united-states) & 5.59 \\
history & 153 & (europe,middle-ages) & 0 & (middle-ages,europe) & 100 \\
history & 141 & (japan,world-war-two) & 0 & (world-war-two,japan) & 100 \\
history & 138 & (military,world-war-two) & 0.72 & (world-war-two,military) & 99.28 \\
history & 136 & (19th-century,united-states) & 0 & (united-states,19th-century) & 100 \\
history & 134 & (soviet-union,world-war-two) & 0.75 & (world-war-two,soviet-union) & 99.25 \\
history & 117 & (20th-century,united-states) & 8.55 & (united-states,20th-century) & 91.45 \\
history & 106 & (ancient-rome,roman-empire) & 84.91 & (roman-empire,ancient-rome) & 15.09 \\
history & 105 & (ancient-history,ancient-rome) & 99.05 & (ancient-rome,ancient-history) & 0.95 \\
 \hline
money  & 3393 & (taxes,united-states) & 0.03 & (united-states,taxes) & 99.97 \\
money & 2087 & (income-tax,united-states) & 0.05 & (united-states,income-tax) & 99.95 \\
 money & 883 & (401k,united-states) & 0 & (united-states,401k) & 100 \\
money & 839 & (income-tax,taxes) & 3.81 & (taxes,income-tax) & 96.19 \\
money & 662 & (tax-deduction,united-states) & 0.15 & (united-states,tax-deduction) & 99.85 \\
money & 638 & (investing,stocks) & 16.3 & (stocks,investing) & 83.7 \\
money & 613 & (ira,united-states) & 0 & (united-states,ira) & 100 \\
money & 604 & (investing,united-states) & 0 & (united-states,investing) & 100 \\
money & 554 & (mortgage,united-states) & 0 & (united-states,mortgage) & 100 \\
money & 541 & (roth-ira,united-states) & 0 & (united-states,roth-ira) & 100 \\
 \hline
 
movies  & 518 & (character,plot-explanation) & 2.9 & (plot-explanation,character) & 97.1 \\
movies & 509 & (marvel-cinematic-universe,plot-explanation) & 0.2 & (plot-explanation,marvel-cinematic-universe) & 99.8 \\
movies & 367 & (game-of-thrones,plot-explanation) & 0.82 & (plot-explanation,game-of-thrones) & 99.18 \\
movies & 242 & (analysis,plot-explanation) & 7.85 & (plot-explanation,analysis) & 92.15 \\
movies & 233 & (avengers-infinity-war,marvel-cinematic-universe) & 0 & (marvel-cinematic-universe,avengers-infinity-war) & 100 \\
movies & 205 & (character,marvel-cinematic-universe) & 100 & (marvel-cinematic-universe,character) & 0 \\
movies & 199 & (avengers-endgame,marvel-cinematic-universe) & 0 & (marvel-cinematic-universe,avengers-endgame) & 100 \\
movies & 184 & (analysis,character) & 29.89 & (character,analysis) & 70.11 \\
movies & 179 & (dialogue,plot-explanation) & 5.59 & (plot-explanation,dialogue) & 94.41 \\
movies & 143 & (ending,plot-explanation) & 2.8 & (plot-explanation,ending) & 97.2 \\
 \hline
 
 music & 519 & (chords,theory) & 0 & (theory,chords) & 100 \\
music & 490 & (chord-theory,chords) & 0 & (chords,chord-theory) & 100 \\
 music & 435 & (harmony,theory) & 0 & (theory,harmony) & 100 \\
music & 410 & (scales,theory) & 0 & (theory,scales) & 100 \\
music & 404 & (chord-theory,theory) & 0 & (theory,chord-theory) & 100 \\
music & 363 & (electric-guitar,guitar) & 0 & (guitar,electric-guitar) & 100 \\
music & 337 & (notation,sheet-music) & 99.41 & (sheet-music,notation) & 0.59 \\
music & 329 & (chords,guitar) & 0 & (guitar,chords) & 100 \\
music & 328 & (chord-progressions,theory) & 0 & (theory,chord-progressions) & 100 \\
music & 306 & (notation,piano) & 0 & (piano,notation) & 100 \\
 \hline
 philosophy & 272 & (logic,philosophy-of-mathematics) & 100 & (philosophy-of-mathematics,logic) & 0 \\
philosophy & 266 & (epistemology,philosophy-of-science) & 94.36 & (philosophy-of-science,epistemology) & 5.64 \\
philosophy & 246 & (fallacies,logic) & 0.41 & (logic,fallacies) & 99.59 \\
philosophy & 193 & (logic,symbolic-logic) & 100 & (symbolic-logic,logic) & 0 \\
philosophy & 186 & (metaphysics,ontology) & 100 & (ontology,metaphysics) & 0 \\
philosophy & 186 & (logic,philosophy-of-logic) & 100 & (philosophy-of-logic,logic) & 0 \\
philosophy & 183 & (argumentation,logic) & 0.55 & (logic,argumentation) & 99.45 \\
philosophy & 179 & (epistemology,metaphysics) & 100 & (metaphysics,epistemology) & 0 \\
philosophy & 179 & (epistemology,logic) & 1.68 & (logic,epistemology) & 98.32 \\
 philosophy & 178 & (logic,proof) & 100 & (proof,logic) & 0 \\
 \hline
 
physics  & 4182 & (homework-and-exercises,newtonian-mechanics) & 99.74 & (newtonian-mechanics,homework-and-exercises) & 0.26 \\
physics & 3658 & (forces,newtonian-mechanics) & 0.52 & (newtonian-mechanics,forces) & 99.48 \\
physics & 2565 & (hilbert-space,quantum-mechanics) & 0 & (quantum-mechanics,hilbert-space) & 100 \\
physics & 2360 & (operators,quantum-mechanics) & 0 & (quantum-mechanics,operators) & 100 \\
physics & 2337 & (quantum-mechanics,wavefunction) & 100 & (wavefunction,quantum-mechanics) & 0 \\
physics & 2238 & (electromagnetism,magnetic-fields) & 99.82 & (magnetic-fields,electromagnetism) & 0.18 \\
physics & 2196 & (homework-and-exercises,quantum-mechanics) & 0 & (quantum-mechanics,homework-and-exercises) & 100 \\
physics & 1988 & (newtonian-gravity,newtonian-mechanics) & 0 & (newtonian-mechanics,newtonian-gravity) & 100 \\
physics & 1767 & (quantum-mechanics,schroedinger-equation) & 100 & (schroedinger-equation,quantum-mechanics) & 0 \\
physics & 1704 & (black-holes,general-relativity) & 0 & (general-relativity,black-holes) & 100 \\

\bottomrule
\end{tabular}%
}
\end{table*}

\begin{table*}[]
\caption{Tag Ordering Statistics - Last 5 domains}
\label{tab:app_tagorder_3}
\small
\resizebox{0.9\textwidth}{!}{%
\begin{tabular}{@{}lrlrlr@{}}
\toprule
Domain & Total & Order-1 & \% & Order-2 & \% \\ \midrule

politics & 570 & (donald-trump,united-states) & 0 & (united-states,donald-trump) & 100 \\
politics & 557 & (president,united-states) & 0 & (united-states,president) & 100 \\
politics & 523 & (presidential-election,united-states) & 0 & (united-states,presidential-election) & 100 \\
politics & 478 & (congress,united-states) & 0 & (united-states,congress) & 100 \\
politics & 475 & (election,united-states) & 0.63 & (united-states,election) & 99.37 \\
politics & 467 & (brexit,united-kingdom) & 0 & (united-kingdom,brexit) & 100 \\
politics & 328 & (constitution,united-states) & 0 & (united-states,constitution) & 100 \\
politics & 282 & (law,united-states) & 0.35 & (united-states,law) & 99.65 \\
politics & 279 & (senate,united-states) & 0.36 & (united-states,senate) & 99.64 \\
politics & 254 & (united-states,voting) & 100 & (voting,united-states) & 0 \\
\hline

rpg & 5330 & (dnd-5e,spells) & 99.21 & (spells,dnd-5e) & 0.79 \\
rpg & 1367 & (dnd-5e,magic-items) & 100 & (magic-items,dnd-5e) & 0 \\
rpg & 1212 & (class-feature,dnd-5e) & 0 & (dnd-5e,class-feature) & 100 \\
rpg & 1204 & (dnd-5e,monsters) & 99.83 & (monsters,dnd-5e) & 0.17 \\
rpg & 1188 & (pathfinder-1e,spells) & 90.24 & (spells,pathfinder-1e) & 9.76 \\
rpg & 959 & (dnd-3.5e,spells) & 72.78 & (spells,dnd-3.5e) & 27.22 \\
rpg & 676 & (dnd-5e,feats) & 99.85 & (feats,dnd-5e) & 0.15 \\
rpg & 632 & (dnd-5e,warlock) & 100 & (warlock,dnd-5e) & 0 \\
rpg & 607 & (balance,dnd-5e) & 0.16 & (dnd-5e,balance) & 99.84 \\
rpg & 567 & (combat,dnd-5e) & 0.53 & (dnd-5e,combat) & 99.47 \\
\hline
 
scifi  & 3514 & (short-stories,story-identification) & 1.05 & (story-identification,short-stories) & 98.95 \\
scifi & 2109 & (marvel,marvel-cinematic-universe) & 76.67 & (marvel-cinematic-universe,marvel) & 23.33 \\
scifi & 2029 & (books,story-identification) & 0.74 & (story-identification,books) & 99.26 \\
scifi & 1922 & (the-lord-of-the-rings,tolkiens-legendarium) & 52.76 & (tolkiens-legendarium,the-lord-of-the-rings) & 47.24 \\
scifi & 1859 & (novel,story-identification) & 1.02 & (story-identification,novel) & 98.98 \\
scifi & 1638 & (movie,story-identification) & 1.47 & (story-identification,movie) & 98.53 \\
scifi & 1497 & (star-trek,star-trek-tng) & 99.67 & (star-trek-tng,star-trek) & 0.33 \\
scifi & 1077 & (aliens,story-identification) & 2.04 & (story-identification,aliens) & 97.96 \\
scifi & 866 & (a-song-of-ice-and-fire,game-of-thrones) & 6.24 & (game-of-thrones,a-song-of-ice-and-fire) & 93.76 \\
scifi & 723 & (star-wars,star-wars-legends) & 100 & (star-wars-legends,star-wars) & 0 \\
\hline
 
 serverfault & 3261 & (linux,ubuntu) & 98.13 & (ubuntu,linux) & 1.87 \\
serverfault & 2865 & (centos,linux) & 1.33 & (linux,centos) & 98.67 \\
serverfault & 2498 & (amazon-ec2,amazon-web-services) & 76.7 & (amazon-web-services,amazon-ec2) & 23.3 \\
serverfault & 2452 & (linux,networking) & 99.14 & (networking,linux) & 0.86 \\
serverfault & 1912 & (apache-2.2,php) & 86.72 & (php,apache-2.2) & 13.28 \\
serverfault & 1803 & (debian,linux) & 1.5 & (linux,debian) & 98.5 \\
serverfault & 1716 & (linux,ssh) & 98.19 & (ssh,linux) & 1.81 \\
serverfault & 1643 & (apache-2.2,linux) & 2.01 & (linux,apache-2.2) & 97.99 \\
serverfault & 1560 & (iptables,linux) & 1.15 & (linux,iptables) & 98.85 \\
serverfault & 1466 & (apache-2.2,virtualhost) & 96.18 & (virtualhost,apache-2.2) & 3.82 \\
\hline
 
travel & 2181 & (uk,visas) & 0.05 & (visas,uk) & 99.95 \\
travel & 1779 & (schengen,visas) & 0.06 & (visas,schengen) & 99.94 \\
travel & 1340 & (usa,visas) & 2.24 & (visas,usa) & 97.76 \\
travel & 871 & (customs-and-immigration,usa) & 0 & (usa,customs-and-immigration) & 100 \\
travel & 795 & (indian-citizens,visas) & 0 & (visas,indian-citizens) & 100 \\
travel & 727 & (transit,visas) & 0 & (visas,transit) & 100 \\
travel & 726 & (customs-and-immigration,visas) & 0 & (visas,customs-and-immigration) & 100 \\
travel & 643 & (standard-visitor-visas,uk) & 0 & (uk,standard-visitor-visas) & 100 \\
travel & 566 & (uk,visa-refusals) & 100 & (visa-refusals,uk) & 0 \\
travel & 511 & (visa-refusals,visas) & 0 & (visas,visa-refusals) & 100 \\
 
\bottomrule
\end{tabular}%
}
\end{table*}

\begin{table}[]
\caption{Tag-Post Overlap:\% posts where at least one tag appears in user texts. EMS\%: single word tag exact match, EMM\%: single \& multi-word tag exact match}
\label{tab:apptag_post_overlap}
\centering
\setlength{\tabcolsep}{10pt}
\resizebox{\columnwidth}{!}{%
\begin{tabular}{l|rr|rr|rr}
\toprule
\multirow{2}{*}{Domain} &
  \multicolumn{2}{c}{Title} &
  \multicolumn{2}{c}{Title+Body} &
  \multicolumn{2}{c}{Title+Body+Answer} \\\cline{2-7}
 &
  EMS &
  EMM &
  EMS &
  EMM &
  EMS &
  EMM \\ \midrule
askubuntu & \blue{56.94} & \blue{71.64} & \blue{77.29} & \blue{88.67} & 81.46 & 91.15 \\ 
aviation & 29.09 & 49.63 & 47.53 & 66.11 & 58.98 & 75.76 \\ 
biology & 17.95 & 29.68 & 33.70 & 47.17 & 42.92 & 56.97 \\ 
chemistry & 19.72 & \textbf{29.26} & 32.81 & \textbf{46.44} & 40.99 & \textbf{56.20} \\ 
cooking & \blue{53.51} & \blue{71.04} & \blue{72.75} & \blue{82.92} & 80.87 & 88.40 \\ 
electronics & \blue{55.24} & \blue{71.11} & \blue{75.21} & \blue{86.69} & 80.6 & 89.99 \\ 
history & 22.23 & 44.34 & 43.76 & 67.23 & 59.66 & 79.51 \\ 
money & 34.47 & \blue{57.89} & 56.32 & \blue{78.66} & 66.62 & 86.23 \\ 
movies & \textbf{9.49} & 34.51 & \textbf{17.69} & \blue{72.92} & \textbf{24.06} & 77.32 \\ 
music & 37.42 & \blue{53.76} & 60.81 & \blue{75.32} & 75.02 & 85.78 \\ 
philosophy & 23.85 & 43.24 & 44.85 & 63.77 & 59.81 & 75.54 \\ 
physics & 24.34 & 40.95 & 40.25 & 61.48 & 48.24 & 70.20 \\ 
politics & 28.37 & \blue{54.11} & 52.64 & \blue{77.52} & 67.41 & 88.21 \\ 
rpg & 23.69 & 41.14 & 44.72 & 65.13 & 57.52 & 76.20 \\ 
scifi & 17.76 & 38.31 & 28.38 & 60.49 & 34.63 & 70.64 \\ 
serverfault & \blue{58.67} & \blue{74.50} & \blue{76.70} & \blue{89.34} & 80.02 & 91.51 \\ 
travel & 45.93 & \blue{63.66} & 65.88 &\blue{ 79.76} & 75.11 & 86.38 \\ 
 \bottomrule
\end{tabular}%
}
\end{table}

\section{Tag-Post Overlap: Full Table}
Table \ref{tab:apptag_post_overlap} shows the tag-post overlap in tabular form similar to Figure \ref{fig:tag_post_overlap} in Section \ref{sec:tag_post_ovrlap}.

\section{Decoding Phase of the MRPG Model}
\label{sec:app_decoding}
We allow the model to generate the tags based on the input parameter maximum output length and then use few heuristics to filter out appropriate tag-tokens and choose the top-k tags. Our heuristics are based on  prior knowledge about how a tag token should be like (1) a tag cannot start or end with a '-' (2) skip the punctuation tokens (3) ignoring adjacent repeated tags.
We then combine the tag tokens between two \tagsep tokens to form the final tag. We also calculate the top-k ($k=1\dots5$) most probable tags based on the combined probability scores of the tag-tokens.

\section{Feature-based Model Configurations:}
For building both the tf-idf and bag of words features we consider unigram and bigram features with a minimum document frequency of 0.00009. We generate  200,000 maximum features. We consider log loss and search hyper-parameter space using alpha = [0.0001,0.001,0.00001] and penalty=[$l_1$, $l_2$] for the Stochastic Gradient Descent One versus rest classifier. For both the models, we find that $l_2$ penalty with 0.00001 alpha yields the best performance.

\section{P-values for Hit@5}
\label{sec:app_pvalues}
Table \ref{tab:app_pvalues} shows the p-values when MRPG model's Hit@5 is compared with MP model. The significance test has been done by one-sided Wilcoxon Test\cite{hollander2013nonparametric}. For k=1,2,3,4 MRPG model's Hit@k shows significant improvements over MP model. MRPG model outperforms all other baselines significantly in Hit@k metrics for each value of k.

\begin{table}[]
\caption{MRPG vs MP: P-values for Hit@5 calculated by one-sided wilcoxon test \cite{hollander2013nonparametric}. The improvement of MRPG is considered significant if it is less than 0.05}
\label{tab:app_pvalues}
\tiny
\centering
\resizebox{0.7\columnwidth}{!}{%
\begin{tabular}{@{}lcc@{}}
\toprule
Domains & P-Values & Is Significant \\ \midrule
askubuntu  &  0.03125  &  Yes \\
aviation  &  0.15625  &  No \\
biology  &  1.00000  &  No \\
chemistry  &  0.03125  &  Yes \\
cooking  &  0.03125  &  Yes \\
electronics  &  0.03125  &  Yes \\
history  &  0.09375  &  No \\
money  &  0.03125  &  Yes \\
movies  &  0.40625  &  No \\
music  &  0.03125  &  Yes \\
philosophy  &  0.50000  &  No \\
physics  &  0.03125  &  Yes \\
politics  &  0.03125  &  Yes \\
rpg  &  0.03125  &  Yes \\
scifi  &  0.03125  &  Yes \\
serverfault  &  0.03125  &  Yes \\
travel  &  0.03125  &  Yes \\
\bottomrule
\end{tabular}%
}
\end{table}

\section{Detailed Tag-Post Coverage \%}
Table \ref{tab:app_tag_post_coverage} shows detailed tag-post coverage.

\begin{table}[]
\caption{Top-n Tag's Post Coverage. \#T:\#distinct tags, 100T\%:Top-100 tag percent among whole tag-space.}
\label{tab:app_tag_post_coverage}
\resizebox{0.9\columnwidth}{!}{%
\begin{tabular}{@{}l|r|rrrr|rr|r@{}}
\toprule
\multicolumn{1}{l}{Domain} &
  \multicolumn{1}{l}{\#T} &
  \multicolumn{1}{l}{Top1} &
  \multicolumn{1}{l}{Top3} &
  \multicolumn{1}{l}{Top5} &
  \multicolumn{1}{l}{Top10} &
  \multicolumn{1}{l}{Top50} &
  \multicolumn{1}{l}{Top100} &
  \multicolumn{1}{l}{100T\%} \\ \midrule
 askubuntu & 3121 & 5.67 & 15.87 & 24.81 & 40.21 & 71.84 & 82.68 & 3.2 \\
aviation & 1002 & 11.05 & 25.81 & 33.87 & 45.93 & 79.13 & 89.43 & 9.98 \\
biology & 739 & 9.22 & 23.91 & 37.84 & 55.05 & 84.39 & 91.76 & 13.53 \\
chemistry & 375 & 23.05 & 42.61 & 48.62 & 61.38 & 87.69 & 95.35 & 26.67 \\
cooking & 833 & 9.55 & 22.45 & 29.55 & 38.99 & 71.45 & 85.19 & 12 \\
electronics & 2226 & 4.94 & 13.84 & 20.88 & 32.81 & 68.96 & 81.98 & 4.49 \\
history & 813 & 10.86 & 25.08 & 35.27 & 45.91 & 80.82 & 89.95 & 12.3 \\
money & 995 & 37.04 & 49.69 & 56.62 & 68.52 & 88.33 & 94.18 & 10.05 \\
movies & 4348 & 36.93 & 49.59 & 56.36 & 66.84 & 81.59 & 85.88 & 2.3 \\
music & 512 & 14.93 & 39.08 & 47.59 & 58.04 & 87.42 & 94.54 & 19.53 \\
philosophy & 559 & 19.39 & 37.1 & 48.56 & 63.3 & 87.29 & 93.77 & 17.89 \\
physics & 893 & 12.7 & 28.35 & 39.99 & 55.1 & 83.98 & 91.68 & 11.2 \\
politics & 739 & 46 & 59.16 & 63.64 & 66.41 & 89.63 & 94.95 & 13.53 \\
rpg & 1195 & 42.5 & 61.23 & 76.9 & 79.75 & 88.01 & 92.66 & 8.37 \\
scifi & 3433 & 27.86 & 47.75 & 62.03 & 70.67 & 81.32 & 85.04 & 2.91 \\
serverfault & 3814 & 11.92 & 22.16 & 29.97 & 42.76 & 72.8 & 82.86 & 2.62 \\
travel & 1891 & 22.2 & 36.03 & 48.34 & 58.34 & 84.39 & 92.36 & 5.29 \\
\bottomrule
\end{tabular}%
}
\end{table}

\section{Effect of Using Answers}
\label{sec:app_using_answers}
We can use answers in those domains or organizations where we already have some answers posted and the tag-prediction approach is being deployed later. The motivation for using answers directly comes from our Tag-Post Overlap analysis in Table \ref{fig:tag_post_overlap}, where we can find a minimum overlap of tags in 70\% of posts in 16/17 domains with the exception of chemistry and biology domains. In these two domains, the overlap increases by around 9-10\%. In some domains, the overlap also increases to 91\%.

\begin{table*}[]
\caption{Effect of Vocabulary Size Reduction on Individual Models Hit@5 Metric}
\label{tab:app_vocab_diff_model_comp}
\tiny
\resizebox{0.6\textwidth}{!}{%
\begin{tabular}{@{}l|rr|rr@{}}
\toprule
\multicolumn{1}{c}{\multirow{2}{*}{Domain}} & \multicolumn{2}{c}{MP}                           & \multicolumn{2}{c}{MRPG}                         \\ \cmidrule(l){2-5}
\multicolumn{1}{c}{}                        & \multicolumn{1}{c}{90} & \multicolumn{1}{c}{85} & \multicolumn{1}{c}{90} & \multicolumn{1}{c}{85} \\ 
\midrule
askubuntu & 80.42 & 75.73 $\Delta$-4.69 & 83.18 & 80.92 $\Delta$-2.26\\
aviation & 77.12 & 73.21 $\Delta$-3.91 & 77.64 & 77.68 $\Delta$0.04\\
biology & 79.31 & 76.35 $\Delta$-2.96 & 78.03 & 77.41 $\Delta$-0.62\\
chemistry & 77.77 & 75.62 $\Delta$-2.15 & 79.51 & 79.63 $\Delta$0.12\\
cooking & 80.42 & 76.81 $\Delta$-3.61 & 85.38 & 85.29 $\Delta$-0.09\\
electronics & 77.92 & 73.69 $\Delta$-4.23 & 81.62 & 80.56 $\Delta$-1.06\\
history & 80.57 & 77.59 $\Delta$-2.98 & 82.29 & 81.21 $\Delta$-1.08\\
money & 84.46 & 80.38 $\Delta$-4.08 & 88.19 & 87.9 $\Delta$-0.29\\
movies & 83.54 & 78.6 $\Delta$-4.94 & 82.77 & 82.8 $\Delta$0.03\\
music & 82.72 & 78.73 $\Delta$-3.99 & 84.37 & 84.18 $\Delta$-0.19\\
philosophy & 79.17 & 74.4 $\Delta$-4.77 & 79.58 & 79.1 $\Delta$-0.48\\
physics & 81.49 & 77.3 $\Delta$-4.19 & 86.48 & 85.78 $\Delta$-0.7\\
politics & 86.43 & 82.4 $\Delta$-4.03 & 91.38 & 90.74 $\Delta$-0.64\\
rpg & 83.71 & 79.41 $\Delta$-4.3 & 89.23 & 88.1 $\Delta$-1.13\\
scifi & 85.81 & 82.22 $\Delta$-3.59 & 91.55 & 90.72 $\Delta$-0.83\\
serverfault & 81.87 & 77.26 $\Delta$-4.61 & 85.9 & 85.04 $\Delta$-0.86\\
travel & 84.09 & 79.41 $\Delta$-4.68 & 89.47 & 88.53 $\Delta$-0.94\\\bottomrule
\end{tabular}%
}
\end{table*}

\begin{table*}[]
\caption{Head Contributions for 90\%, 85\% and 95\% Post Coverage by Tags: P represents total correct predictions \% by P-Head only and G represents total correct prediction \% by G-Head only}
\label{tab:head_contrib}
\tiny
\resizebox{0.8\textwidth}{!}{%
\begin{tabular}{@{}l|rrrrrr@{}}
\toprule
\multicolumn{1}{c}{\multirow{2}{*}{Domain}} &
  \multicolumn{2}{c}{90} &
  \multicolumn{2}{c}{85} &
  \multicolumn{2}{c}{95} \\ \cmidrule(l){2-7} 
\multicolumn{1}{c}{} &
  \multicolumn{1}{c}{P} &
  \multicolumn{1}{c}{G} &
  \multicolumn{1}{c}{P} &
  \multicolumn{1}{c}{G} &
  \multicolumn{1}{c}{P} &
  \multicolumn{1}{c}{G} \\ \midrule
  askubuntu & 49.66 & 11.85 & 46.5(-3.16) & 14.3(2.45) & 59.27(9.61) & 7.54(-4.31) \\
aviation & 53.8 & 10.42 & 47.78(-6.02) & 13.76(3.34) & 61.54(7.74) & 5.41(-5.01) \\
biology & 53.86 & 9.88 & 49.32(-4.54) & 11.47(1.59) & 61.1(7.24) & 5.8(-4.08) \\
chemistry & 55.18 & 10.61 & 50.89(-4.29) & 12.96(2.35) & 63.87(8.69) & 6.31(-4.3) \\
cooking & 55.03 & 10.95 & 47.15(-7.88) & 14.93(3.98) & 64.45(9.42) & 6.24(-4.71) \\
electronics & 52.07 & 11.28 & 46.28(-5.79) & 14.39(3.11) & 59.52(7.45) & 7.24(-4.04) \\
history & 55.06 & 9.91 & 52.15(-2.91) & 10.51(0.6) & 65.25(10.19) & 3.94(-5.97) \\
money & 50.17 & 10.23 & 43.74(-6.43) & 12.82(2.59) & 61.49(11.32) & 5.99(-4.24) \\
movies & 68.51 & 4.55 & 58.63(-9.88) & 7.4(2.85) & 74.17(5.66) & 1.61(-2.94) \\
music & 56.46 & 10.18 & 50.32(-6.14) & 13.19(3.01) & 64.44(7.98) & 6.69(-3.49) \\
philosophy & 58.98 & 7.81 & 53.02(-5.96) & 10.85(3.04) & 64.9(5.92) & 4.61(-3.2) \\
physics & 45.01 & 12 & 39.84(-5.17) & 14.73(2.73) & 55.04(10.03) & 8.41(-3.59) \\
politics & 52.6 & 9.14 & 44.87(-7.73) & 12.81(3.67) & 67.38(14.78) & 4.51(-4.63) \\
rpg & 51.89 & 7.68 & 39.03(-12.86) & 11.34(3.66) & 68.68(16.79) & 3.97(-3.71) \\
scifi & 74.64 & 5.11 & 62.79(-11.85) & 8.37(3.26) & 84.69(10.05) & 1.6(-3.51) \\
serverfault & 50.63 & 11.9 & 42.96(-7.67) & 15.97(4.07) & 58.55(7.92) & 8.45(-3.45) \\
travel & 46.26 & 12.39 & 40.6(-5.66) & 15.94(3.55) & 53.59(7.33) & 8.19(-4.2) \\
\bottomrule
\end{tabular}%
}
\end{table*}

\begin{table*}[]
\caption{Model Performance (Hit@k) for MP and MRPG models}
\label{tab:app_model_comp}
\resizebox{0.85\textwidth}{!}{%
\begin{tabular}{@{}l|rrrrr|rrrrr@{}}
\toprule
\multicolumn{1}{c}{\multirow{2}{*}{Domain}} & \multicolumn{5}{c}{MP} & \multicolumn{5}{c}{MRPG} \\ \cmidrule(l){2-11} 
\multicolumn{1}{c}{} &
  \multicolumn{1}{c}{Hit@1} &
  \multicolumn{1}{c}{Hit@2} &
  \multicolumn{1}{c}{Hit@3} &
  \multicolumn{1}{c}{Hit@4} &
  \multicolumn{1}{c}{Hit@5} &
  \multicolumn{1}{c}{Hit@1} &
  \multicolumn{1}{c}{Hit@2} &
  \multicolumn{1}{c}{Hit@3} &
  \multicolumn{1}{c}{Hit@4} &
  \multicolumn{1}{c}{Hit@5} \\ \midrule
  askubuntu & 31.59 $\pm$ 0.14 & 50.89 $\pm$ 64.85 & 0.21 $\pm$ 0.14 & 74.23 $\pm$ 0.31 & 80.44 $\pm$ 0.11 & 50.86 $\pm$ 0.13 & 72.72  $\pm$ 0.1 & 80.19  $\pm$ 0.09 & 81.71  $\pm$ 0.17 & 82.94  $\pm$ 0.15 \\ 
aviation & 30.72 $\pm$ 0.95 & 48.39 $\pm$ 62.05 & 0.93 $\pm$ 0.55 & 72.23 $\pm$ 0.66 & 77.09 $\pm$ 0.44 & 47.28 $\pm$ 0.19 & 67.37  $\pm$ 0.36 & 75.21  $\pm$ 0.47 & 76.02  $\pm$ 0.45 & 77.63  $\pm$ 0.56 \\ 
biology & 34.66 $\pm$ 0.64 & 50.81 $\pm$ 63.77 & 1.01 $\pm$ 0.31 & 73.96 $\pm$ 0.32 & 78.96 $\pm$ 0.34 & 49.67 $\pm$ 0.7 & 68.8  $\pm$ 0.37 & 75.7  $\pm$ 0.42 & 76.51  $\pm$ 0.44 & 77.55  $\pm$ 0.41 \\ 
chemistry & 38.83 $\pm$ 0.51 & 54.77 $\pm$ 65.84 & 0.36 $\pm$ 0.72 & 73.35 $\pm$ 0.34 & 77.66 $\pm$ 0.1 & 50.28 $\pm$ 0.15 & 69.81  $\pm$ 0.33 & 76.43  $\pm$ 0.3 & 77.23  $\pm$ 0.38 & 79.17  $\pm$ 0.45 \\ 
cooking & 35.46 $\pm$ 0.28 & 56.17 $\pm$ 67.2 & 0.79 $\pm$ 0.82 & 76.21 $\pm$ 0.95 & 80.86 $\pm$ 0.42 & 52.43 $\pm$ 0.68 & 75.61  $\pm$ 0.15 & 82.73  $\pm$ 0.25 & 83.74  $\pm$ 0.31 & 85.18  $\pm$ 0.29 \\ 
electronics & 28.67 $\pm$ 0.43 & 47.28 $\pm$ 61.78 & 0.73 $\pm$ 0.54 & 70.97 $\pm$ 0.24 & 77.51 $\pm$ 0.26 & 49.64 $\pm$ 0.63 & 71.4  $\pm$ 0.53 & 78.92  $\pm$ 0.46 & 80.06  $\pm$ 0.47 & 81.3  $\pm$ 0.53 \\ 
history & 34.18 $\pm$ 1.21 & 54.16 $\pm$ 66.47 & 1.24 $\pm$ 0.97 & 76.22 $\pm$ 0.53 & 80.45 $\pm$ 0.09 & 54.07 $\pm$ 0.1 & 73.56  $\pm$ 0.62 & 79.5  $\pm$ 0.82 & 80.29  $\pm$ 0.88 & 81.23  $\pm$ 1 \\ 
money & 51.05 $\pm$ 0.44 & 66.28 $\pm$ 75.43 & 0.98 $\pm$ 0.43 & 81.01 $\pm$ 0.24 & 84.15 $\pm$ 0.23 & 60.59 $\pm$ 0.47 & 78.89  $\pm$ 0.14 & 86.01  $\pm$ 0.4 & 86.75  $\pm$ 0.41 & 87.94  $\pm$ 0.42 \\ 
movies & 50.06 $\pm$ 0.5 & 64.28 $\pm$ 73.58 & 1.53 $\pm$ 0.88 & 79.48 $\pm$ 0.7 & 82.91 $\pm$ 0.55 & 57.33 $\pm$ 0.44 & 78.41  $\pm$ 0.28 & 82.03  $\pm$ 0.69 & 83.05  $\pm$ 0.9 & 83.25  $\pm$ 0.99 \\ 
music & 37.03 $\pm$ 0.41 & 57.17 $\pm$ 68.79 & 0.62 $\pm$ 0.27 & 77.76 $\pm$ 0.43 & 82.66 $\pm$ 0.26 & 53.17 $\pm$ 0.69 & 75.17  $\pm$ 0.35 & 81.46  $\pm$ 0.52 & 82.28  $\pm$ 0.49 & 83.71  $\pm$ 0.51 \\ 
philosophy & 34.9 $\pm$ 0.8 & 52.94 $\pm$ 66.03 & 0.26 $\pm$ 0.77 & 75.46 $\pm$ 0.53 & 79.45 $\pm$ 0.2 & 53.09 $\pm$ 0.95 & 72.01  $\pm$ 0.27 & 78.03  $\pm$ 0.59 & 78.76  $\pm$ 0.52 & 79.49  $\pm$ 0.56 \\ 
physics & 41.27 $\pm$ 0.46 & 60.63 $\pm$ 70.59 & 0.49 $\pm$ 0.18 & 77.39 $\pm$ 0.31 & 81.12 $\pm$ 0.22 & 57.96 $\pm$ 0.28 & 75.49  $\pm$ 0.28 & 83.5  $\pm$ 0.38 & 84.47  $\pm$ 0.41 & 86.34  $\pm$ 0.37 \\ 
politics & 65.26 $\pm$ 1.76 & 73.87 $\pm$ 78.43 & 1.18 $\pm$ 0.86 & 84.04 $\pm$ 0.42 & 86.29 $\pm$ 0.25 & 71.61 $\pm$ 0.5 & 82.92  $\pm$ 0.31 & 88.97  $\pm$ 0.46 & 89.91  $\pm$ 0.42 & 90.98  $\pm$ 0.46 \\ 
rpg & 68.68 $\pm$ 0.29 & 75.93 $\pm$ 79.43 & 0.31 $\pm$ 0.37 & 81.66 $\pm$ 0.36 & 83.31 $\pm$ 0.33 & 72.85 $\pm$ 0.4 & 81.91  $\pm$ 0.17 & 87.59  $\pm$ 0.12 & 88.28  $\pm$ 0.16 & 89.09  $\pm$ 0.16 \\ 
scifi & 76.65 $\pm$ 0.34 & 80.64 $\pm$ 82.69 & 0.42 $\pm$ 0.36 & 84.89 $\pm$ 0.15 & 85.91 $\pm$ 0.11 & 81.99 $\pm$ 0.12 & 86.2  $\pm$ 0.21 & 90.41  $\pm$ 0.23 & 90.85  $\pm$ 0.28 & 91.53  $\pm$ 0.32 \\ 
serverfault & 25.9 $\pm$ 0.12 & 50.99 $\pm$ 65.52 & 0.59 $\pm$ 0.25 & 75.24 $\pm$ 0.34 & 81.66 $\pm$ 0.16 & 53.05 $\pm$ 0.38 & 75.36  $\pm$ 0.18 & 82.56  $\pm$ 0.17 & 84.16  $\pm$ 0.29 & 85.82  $\pm$ 0.26 \\ 
travel & 45.04 $\pm$ 0.51 & 63.62 $\pm$ 72.52 & 0.14 $\pm$ 0.41 & 79.96 $\pm$ 0.24 & 83.96 $\pm$ 0.12 & 58.7 $\pm$ 0.34 & 78.57  $\pm$ 0.2 & 86.75  $\pm$ 0.25 & 87.78  $\pm$ 0.28 & 89.5  $\pm$ 0.3 \\ 
  \bottomrule
\end{tabular}%
}
\end{table*}


\end{document}